\documentclass[]{jair}

\setcopyright{cc}
\acmDOI{10.1613/jair.1.20934}

\JAIRAE{Chang Xu}
\JAIRTrack{} 
\acmVolume{86}
\acmArticle{55}
\acmMonth{08}
\acmYear{2026}

\RequirePackage[
  datamodel=acmdatamodel,
  style=acmauthoryear,
  backend=biber,
  giveninits=true,
  uniquename=init
  ]{biblatex}

\usepackage{amsmath}
\usepackage{cleveref}
\usepackage{url}
\usepackage{graphicx}

\usepackage{algorithm}
\usepackage{algorithmic}

\usepackage{tikz}
\usepackage{subfig}

\usetikzlibrary{decorations.pathreplacing,calc}
\newcommand{\tikzmark}[1]{\tikz[overlay,remember picture] \node (#1) {};}
\newcommand*{\AddNote}[4]{%
    \begin{tikzpicture}[overlay, remember picture]
        \draw [decoration={brace,amplitude=0.3em},decorate,ultra thick,black]
            ($(#3)!(#1.north)!($(#3)-(0,1)$)$) --  
            ($(#3)!(#2.south)!($(#3)-(0,1)$)$)
                node [align=center, text width=2.5cm, pos=0.5, anchor=west] {#4};
    \end{tikzpicture}
}%
\usepackage{multirow}
\usepackage{mathtools}
\usepackage{amsmath, amsthm}
\newtheorem{proposition}{Proposition}
\usepackage{booktabs}

\begin{document}

\title[CollaFuse]{CollaFuse: Collaborative Diffusion Models}

\author{Simeon Allmendinger}
\authornote{Both authors contributed equally to this research.}
\authornote{Corresponding author.}
\orcid{0009-0005-8741-7734}
\email{simeon.allmendinger@uni-bayreuth.de}
\affiliation{%
  \institution{University of Bayreuth}
  \city{Bayreuth}
  \country{Germany}
}
\affiliation{%
  \institution{Fraunhofer FIT}
  \city{Bayreuth}
  \country{Germany}
}

\author{Domenique Zipperling}
\authornotemark[1]
\orcid{0009-0003-4598-9051}
\email{domenique.zipperling@uni-bayreuth.de}
\affiliation{%
  \institution{University of Bayreuth}
  \city{Bayreuth}
  \country{Germany}
}
\affiliation{%
  \institution{Fraunhofer FIT}
  \city{Bayreuth}
  \country{Germany}
}

\author{Lukas Struppek}
\orcid{0000-0003-0626-3672}
\email{lukas.struppek@tu-darmstadt.de}
\affiliation{%
  \institution{Technical University of Darmstadt}
  \city{Darmstadt}
  \country{Germany}
}

\author{Niklas Kühl}
\orcid{0000-0001-6750-0876}
\email{kuehl@uni-bayreuth.de}
\affiliation{%
  \institution{University of Bayreuth}
  \city{Bayreuth}
  \country{Germany}
}
\affiliation{%
  \institution{Fraunhofer FIT}
  \city{Bayreuth}
  \country{Germany}
}

\renewcommand{\shortauthors}{Allmendinger, Zipperling, Struppek \& Kühl}

\begin{abstract}
In the landscape of generative artificial intelligence, diffusion-based models have emerged as a promising method for generating synthetic images. However, the application of diffusion models poses numerous challenges, particularly concerning data availability, computational requirements, and privacy. Traditional approaches to address these shortcomings, like federated learning, often impose significant computational burdens on individual clients, especially those with constrained resources. In response to these challenges, we introduce the novel approach \texttt{CollaFuse} for distributed collaborative diffusion models inspired by split learning. Our approach facilitates collaborative training of diffusion models while alleviating client computational burdens during image synthesis. This reduced computational burden is achieved by retaining data and computationally inexpensive processes locally at each client while outsourcing the computationally expensive processes to shared, more efficient server resources. Through experiments on the common datasets CelebA, CIFAR-10, and Animals-with-Attributes2, our approach demonstrates enhanced performance while decreasing information disclosure as it reduces the necessity for sharing raw data. These capabilities hold significant potential across various application areas, including the design of edge computing solutions. Thus, our work advances distributed machine learning by contributing to the evolution of collaborative diffusion models.
\end{abstract}



\received{08 November 2025}
\received[accepted]{18 June 2026}

\maketitle

\section{Introduction}

Generative artificial intelligence (GenAI) methods exhibit astonishing results in generating images, among other modalities like music~\citep{mariani2024multisource} and video~\citep{singer2023makeavideo,videoworldsimulators2024} with promising applications in areas like healthcare~\citep{sun2023aligning} to help reduce biases~\citep{Ramachandranpillai2024}.
Recent computer vision advancements primarily rely on diffusion models~\citep{Ho2020,song20improved} that generate synthetic images from random noise through iterative denoising steps. We formally introduce diffusion models and related work in \cref{sec:background_diffusion}.
In contrast to more traditional approaches~\citep{Kingma2014VAE,goodfellow20gans}, diffusion models excel in providing high sample quality and strong mode coverage~\citep{KAZEROUNI2023102846,nichol21improving,dhariwal21beating}.
However, the strides in GenAI require large amounts of data, and the generation process itself is computationally expensive due to the multiple required denoising steps. 

\begin{figure*}[ht]
    \centering{\includegraphics[width=.8\linewidth]{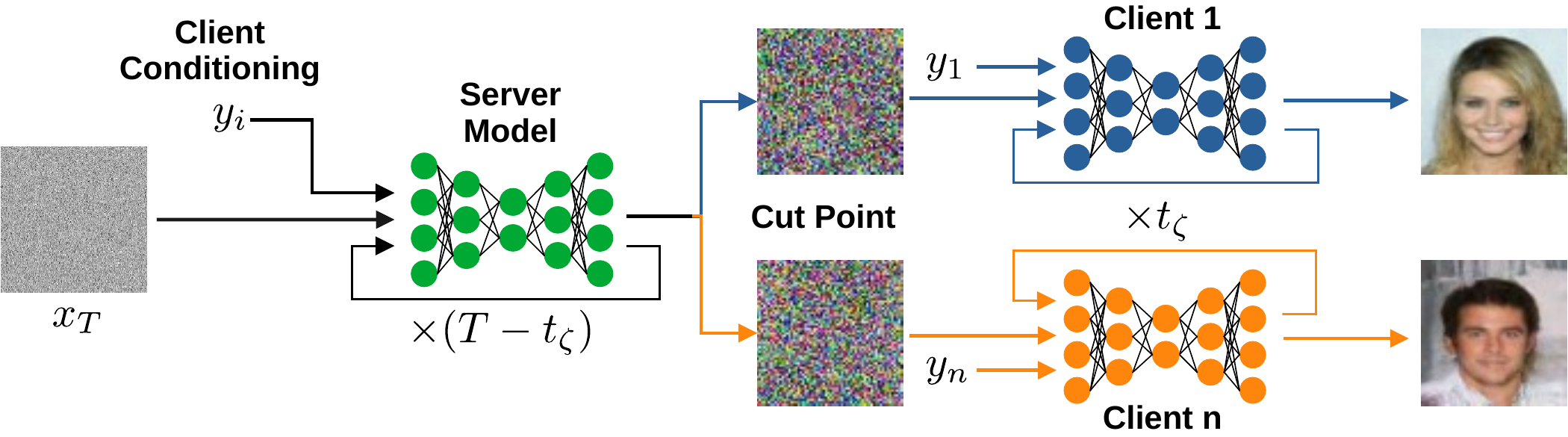}}
    \caption{Overview of CollaFuse for collaborative image synthesis by splitting the denoising process between the server and clients. Based on client-specific conditioning $y_i$, the first $T-t_\zeta$ denoising steps are run on a trusted server model, while the remaining $t_\zeta$ denoising steps are run locally by the $i$-th client. Thereby, external computing resources can be utilized while keeping the clients' raw data private.}
    \label{fig:framework_workflow_schema}
    \Description{
    A horizontal schematic of the CollaFuse pipeline reading left to right. 
    A pure-noise square image labeled x-sub-T feeds into a central neural network drawn as connected nodes, labeled ``Server Model,'' which runs T minus t-zeta denoising steps.
    Above it, client conditioning labels y-sub-i are shown as an input. 
    After a vertical dashed line marked ``Cut Point,'' the partially denoised output branches to two separate client networks: Client 1 (blue nodes) producing a generated female face, and Client n (orange nodes) producing a generated male face, each running t-zeta steps locally.
    Arrows indicate that the server does the heavy initial denoising while each client finishes locally on its own private conditioning, so raw data never leaves the client.
    }
\end{figure*}

While large companies possess the necessary data and computational resources for training diffusion models, smaller organizations and private clients may face challenges in providing the required resources, limiting their ability to train and implement recent GenAI models~\citep{zajac2023}. This might lead to high dependency on a few key players and even prevent the application of such models completely due to local data protection regulations. For example, the popular Stable Diffusion v2 has been trained on 256 A100 GPUs for about 200,000 hours~\citep{Rombach_2022_CVPR}. But even smaller models still set notably high hardware and data requirements.

To address data and resource availability constraints, organizations and clients can join forces to train machine learning models collaboratively with other clients in a decentralized way.
A prominent representative of collaborative training is federated learning~\citep{McMahan2017FL}. Conceptually, each client trains an individual model locally on their private data. After several training steps, the model parameters are sent to the server to build a global model by aggregating the individual weights.
The server distributed this global model back to all clients. As the necessity for each client to train and share an entire model remains, high computational resources for each client are entailed~\citep{SplitFed2022}.
Furthermore, in addition to privacy risk from the generative GenAI models in general~\citep{Hintersdorf2024}, federated learning introduces new potential privacy risks~\citep{shokri17meminf,zhu19leakage}.
As an alternative learning paradigm, split learning~\citep{gupta2018distributed} supports collaborative model training by splitting the model into server-side and client-side components. In the conventional split learning setup, clients share only intermediate network activations with the server side, where the final computations are performed.
Unlike federated learning, split learning reduces the computational load on clients and enhances privacy protection by sharing only intermediate representations instead of raw data or model weights~\citep{Duan2023}.
We provide an overview of collaborative training for generative models in \cref{sec:background_collaborative}.
Our analysis reveals that current approaches rely mostly on federated learning.
By not utilizing split learning with diffusion models, they overlook the privacy and resource benefits of split learning, especially from the clients' perspective.

\noindent\textbf{Our Approach:}
In addressing the challenges posed by the data, computation, and privacy requirements of diffusion-based GenAI, we present our collaborative diffusion models in \cref{sec:approach}.
We introduce the novel collaborative learning and inference approach \texttt{CollaFuse} tailored specifically for diffusion models.
Drawing inspiration from the split learning framework, our approach divides the iterative denoising steps of diffusion models into two components.
The computation of the initial denoising steps is carried out by a shared model on a server, with limited information disclosure due to the inherent noise in the training data and intermediate generated samples. 
Subsequently, the client's model then performs the remaining denoising steps, which are usually significantly fewer than the denoising steps on the server side. \texttt{CollaFuse} also allows for personalized image conditioning by incorporating attribute labels during the generation. 
Our empirical results demonstrate that our collaborative diffusion approach improves the image quality compared to a setting where each client trains its own local diffusion model. By sharing a server model that performs most of the computationally heavy denoising steps, the computational burdens for each client are comparably small.
At the same time, clients can better approximate their individual data distribution, which enables them to generate better characteristic features.
Beyond these performance improvements, we also conduct dedicated privacy analyses, including attribute inference~\citep{fredriskon2014attribute} and inversion attack~\citep{mi_confidence_fredriskon} experiments, to quantify information disclosure at different collaboration cut points.
This allows us to empirically evaluate how collaborative diffusion can balance image fidelity with privacy protection.

The proportion of denoising steps carried out on the server and client sides, respectively, is controlled by a single parameter called the cut point. The higher the cut point, the more steps are computed on the client side. Whereas our approach is in principle also applicable to other diffusion model architectures, we focus our experiments in \cref{sec:experiments} on the common Denoising Diffusion Probabilistic Models (DDPM)~\citep{Ho2020} and Latent Diffusion Models (LDMs)~\citep{Rombach_2022_CVPR}.
In summary, we make the following contributions:

\begin{enumerate}
    \item We introduce \texttt{CollaFuse}, to the best of our knowledge the first split-denoising (timestep-partitioned) collaborative diffusion framework, which consists of a shared server component trained by multiple clients without revealing their original training data to other clients or the server.
    \item \texttt{CollaFuse} allows clients to outsource most of their computationally expensive denoising steps during training and inference to a shared server model.
    \item \texttt{CollaFuse} improves the image quality compared to the setting where each client trains a local diffusion model on its own data.
\end{enumerate}
\section{Background and Related Work}\label{sec:background}
We start by introducing diffusion models for generative image synthesis. We further describe related distributed collaborative machine learning approaches, such as federated and split learning, and their utilization for image synthesis or generative AI more generally.

\subsection{Diffusion Models}\label{sec:background_diffusion}
The Denoising Diffusion Probabilistic Models (DDPMs)~\citep{Ho2020} mark a significant advancement in generative image synthesis consisting of a diffusion and denoising process. The diffusion process is a Markov chain with $T$ timesteps that transforms a training image $x_0$ to a noisy image $x_T$ that follows a random Gaussian distribution. The diffusion process of an image $x_t$ at time step $t$ is mathematically defined as
\begin{equation}
x_t = \sqrt{\alpha_t} x_{t-1} + \sqrt{1 - \alpha_t} \epsilon, \text{ with } t=1,\ldots,T.
\label{eq:diffusion}
\end{equation}
Here, \( \alpha_t \) denotes the variance schedule, and \( \epsilon \) is the added Gaussian noise. A denoising network $\epsilon_\theta(x_t, t)$ with parameters $\theta$ is then trained to reverse the diffusion process and predict the noise added to the sample $x_t$ during time step $t$. Most denoising networks are built upon the common U-Net~\citep{ronneberger15unet} architecture. With the denoising network, the image generation process, which iteratively removes the predicted noise $\epsilon_\theta(x_t, t)$ from the noisy sample $x_t$, can be defined as
\begin{equation}
x_{t-1} = \frac{1}{\sqrt{\alpha_t}} \left(x_t - \frac{1-\alpha_t}{\sqrt{1-\bar{\alpha}_t}} \epsilon_\theta(x_t, t)\right) 
\end{equation}
with $\bar{\alpha}_t = \prod^t_{s=1} \alpha_s$. Based on the idea of Markov chains, the distribution of the intermediate noise predictions in the denoising process $p_\theta(\cdot)$ is defined by 
\begin{equation}
    p_\theta(x_{0:T})=p(x_{T})\cdot\prod^T_{t=1}p_\theta(x_{t-1}|x_{t}) 
    \label{eq:denoising_process}
\end{equation}
with $p(x_{T}) = \mathcal{N}(x_T; \mathbf{0}, \mathbf{I})$. In DDPMs, the iterative application of U-Nets across \( T \) timesteps is a fundamental characteristic, enabling the model to refine noisy data into structured outputs progressively. The Imagen model~\citep{saharia2022photorealistic}, as one of the most recognized text-conditioned diffusion models in the community, builds upon DDPMs. The authors employ a frozen text encoder and dynamic thresholding to generate photorealistic images conditioned by the text prompt $y$. The loss function of DDPMs can be expressed by
\begin{equation}
\mathcal{L}= \mathbb{E}_{\,t\sim\mathcal{U}[1,T],\;\epsilon\sim\mathcal{N}(0,\mathbf{I})}
\big[\, \|\epsilon_\theta(x_t, t, y) - \epsilon\|_2^2 \,\big]
\label{eq:loss_fkt}
\end{equation}
The Imagen model additionally adds a guidance weight \( \omega_t \), which is integral to the denoising process.
This guidance weight modulates the influence of the predicted noise \( \epsilon_\theta(x_t, t, y) \) at each timestep \( t \), enabling precise control over the image generation process, particularly in maintaining fidelity to the target distribution. 
For simplicity, we leave the explicit embedding process out of our notation and implicitly assume that all text labels have already been embedded before feeding the embeddings into the U-Net $\epsilon_\theta(\cdot)$. 

LDMs~\citep{Rombach_2022_CVPR} extend DDPMs to the latent space of image representations, enhancing computational efficiency and scalability~\citep{Rombach_2022_CVPR}. By applying the diffusion process in a lower-dimensional latent space, LDMs generate high-quality images with less computational overhead. Autoencoders encode images into latent representations, where denoising occurs, and then reconstruct the final image.

\subsection{Collaborative Learning in Generative AI}\label{sec:background_collaborative}

Collaborative learning has become an essential paradigm for training models on distributed data while preserving privacy and reducing local computational costs.
It enables multiple clients to jointly optimize model parameters without sharing raw data, forming the basis for methods such as federated and split learning.

\textbf{Collaborative learning:} Federated learning (FL)~\citep{McMahan2017FL} and split learning (SL)~\citep{gupta2018distributed} are among the most prominent approaches for training machine learning models collaboratively on distributed data sources.
FL utilizes distributed data, with clients independently training models on their unique datasets.
These models are subsequently shared, aggregated, and redistributed. The cycle repeats until the models converge.
Conversely, SL divides a model among clients and a central server, decreasing the computational load on clients.
Moreover, clients have the option to use FL for model aggregation, leading to the development of SplitFed learning~\citep{SplitFed2022}.
These techniques have found applications in diverse fields such as the automotive industry~\citep{Schwermer2023FLEV}, energy management~\citep{Schwerner2020FLEngergy}, and healthcare~\citep{Madhura2022}, where they are combined with both discriminative~\citep{Nazir2023} and generative AI approaches~\citep{Shen2023}. 

\textbf{Collaborative learning in generative AI:} Especially for image synthesis, FL and SL possess potential owing to the extensive volume of data involved.
Before diffusion models took over as the predominant architecture for image synthesis, \textit{Generative Adversarial Networks (GANs)}~\citep{goodfellow20gans} represented the most widely used generative architecture.
GANs are composed of two components: a generator for synthesizing images and a discriminator trained to distinguish between real and synthetic images. 
Existing research on collaborative training of GANs demonstrates different ways to integrate these two components within the FL process.
\citet{Hardy2019MD-GAN} introduced FL-GAN, adopting the standard FL procedure for both discriminator and generator. 
This vanilla approach was compared to MD-GAN, where FL is applied only to the discriminator, while the generator is trained directly by a central server.
Building upon this foundation, \citet{fan2020federated} empirically analyzed several strategies for synchronizing discriminators and generators across clients, demonstrating that the best results are achieved when synchronizing both components jointly. 
\citet{Li2022IFL-GAN} improved FL-GANs by employing maximum mean discrepancy for generator updates, while more recent work addresses heterogeneity issues~\citep{Ma2024,Zhao2023}. 
Because FL places substantial strain on computational and communication resources, new methods have been developed to alleviate these burdens.
\citet{Lai2024} proposed an on-demand, quantized, energy-efficient federated diffusion approach to enhance the training efficiency of generative AI models in mobile edge networks by dynamically adjusting quantization, reducing energy and communication consumption while preserving data quality and diversity.

In addition, several studies focus on improving privacy in FL-GANs. 
\citet{augenstein2020generative} proposed a differentially private federated GAN framework, while \citet{Veeraragavan2023SecuringFLGAN} combined consortium blockchains with an efficient secret sharing algorithm to address trust-related weaknesses in existing solutions. 
Beyond FL, GANs have also been collaboratively trained using SL~\citep{Feng.2023IoTSL}. 
Although not explicitly focused on distributed learning, \citet{ohta2023lambdasplit} presented a privacy-preserving SL approach for GANs that could be extended to collaborative learning. 
Other work combines both paradigms, known as SplitFed learning, to leverage the strengths of each~\citep{Kortoi2022,Benshun2023SL-GAN}. 

While these studies demonstrate the feasibility of collaborative learning for GAN-based architectures, their direct transfer to diffusion models remains challenging. 
Diffusion models rely on iterative denoising and stochastic sampling processes that are computationally intensive and require consistent synchronization of parameters over thousands of steps. 
These characteristics complicate straightforward adaptation of existing FL and SL schemes, motivating the development of new collaborative frameworks tailored to diffusion-based generative models. 

\textbf{Collaborative learning of diffusion models:}
In the domain of diffusion models, research on collaborative training methods is still scarce.~\citet{jothiraj2023phoenix} made a first step and introduced the Phoenix technique for training unconditional diffusion models in a horizontal FL setting. Their objective is to address mode coverage issues often seen in distributed datasets that are not independent and identically distributed. Their data-sharing approach boosts performance by sharing only 4-5\% of the data among clients, minimizing communication overhead. Personalization and threshold filtering techniques outperform comparison methods in terms of precision and recall, but fall short in image quality compared to the proposed technique. The paper suggests further exploration to enhance image quality in future work.~\citet{Song2024} proposed enhancing FL with knowledge distillation by dynamically adjusting distillation weights, using data generated by a collaboratively trained diffusion model (FedAvg), thus addressing data heterogeneity and preserving privacy by eliminating the need for public datasets.

Moreover, the potential of FL for AI-generated content, especially for DDPMs, was demonstrated by~\citet{Huang2024FLinAIGC}. The authors discuss three different approaches for diffusion models in FL settings. A parallel approach mimicking the conventional FL. A separate split approach combines FL with SL. As a third solution, the authors discussed a sequential approach in which one client receives the current model from the server, trains the model on its data, and then transmits the current version to the next client.  The trained model returns to the server only after every client has trained the model once. Based on the sequential FL, a LoRA-based~\citep{edward22lora} federated fine-tuning scheme is designed and examined in more detail, demonstrating the advantages of faster convergence time and reduced memory consumption during the tuning process. Further,~\citet{goede2024fltraining} adapt the federated averaging algorithm for training DDPMs.~\citet{mendieta2024navigating} trained diffusion models with one-shot FL under provable privacy budgets with differential privacy.

By mainly focusing on FL for GANs~\citep{Little2023}, current literature has neglected the benefits of different collaborative paradigms and GenAI architectures so far. Combining DDPMs and SL promises various benefits, including reducing local resource requirements and increasing data privacy. \citet{Huang2024FLinAIGC} is the only identified study utilizing SL for diffusion models. However, it combines SL with FL, which increases privacy risks compared to a standard SL approach, as client models are still shared. Moreover, it overlooks the impact of model split configuration on factors such as performance. Our proposed approach for distributed collaborative image synthesis with diffusion models taps into these advantages and combines the research areas of diffusion models and collaborative learning. Focusing exclusively on split learning mitigates privacy risks by sharing only intermediate results.
Additionally, we treat the timestep split configuration (the cut point determining how many denoising steps are performed client-side) as a hyperparameter and analyze its impact.
We emphasize that CollaFuse is inspired by split learning but does not split a network by layers or activations; instead it partitions the reverse denoising trajectory across timesteps, so the privacy and communication intuitions of conventional split learning do not transfer directly.

\section{Collaborative Diffusion Models}\label{sec:approach}

We now formally introduce our novel approach \texttt{CollaFuse} for enabling collaborative image generation with diffusion models. In our setting, a certain number $k$ of clients $c \in C=\{c_1, c_2, ...,c_k\} $ wants to collaboratively train a diffusion model for image synthesis. Although we assume that each client has a dataset from a similar domain, e.g., facial images, the specific feature distribution may differ.
To stay with the facial image example, client \texttt{A} may have a dataset of facial images with eyeglasses, whereas client \texttt{B}'s dataset consists only of faces without eyeglasses. All clients now want to train a shared U-Net $\epsilon_\theta^s$ on the server that is available to each client and computes the initial denoising steps.
Additionally, each client $c \in C$ trains an individual U-Net $\epsilon_\theta^c$ that is maintained locally and computes solely the remaining denoising steps.
For notation simplicity, we assume that $\theta$ denotes the weights of each individual model, so there exist no shared weights between the client models and the server model. 

The computational split between server and clients is manually set by the cut point $t_\zeta \in [0,T]$ that specifies the number of denoising steps performed on the client side after $T-t_\zeta$ steps were computed by the shared server model. The cut point is set as a hyperparameter and kept fixed during training and inference. For $t_\zeta=0$, all denoising steps are computed by the server, which is trained on the joint set of all clients' data.
For $t_\zeta=T$, each client trains an individual diffusion model on its data that performs all denoising steps without any shared server model.
The approximated data distribution of our collaborative denoising approach is formalized in \Cref{eq:split_denoising_process} with $p(x_{T})=\mathcal{N}(x_T;0,\mathbf{I})$:
\begin{equation}
    p_{\theta^s,\theta^c}(x_{0:T}) = p(x_T)
    \prod_{t=t_\zeta+1}^{T} p_{\theta^s}(x_{t-1}\mid x_t)\cdot
    \prod_{t=1}^{t_\zeta} p_{\theta^c}(x_{t-1}\mid x_t)
    \label{eq:split_denoising_process} 
\end{equation}
Here, the first product operator describes the distribution approximated by the server model with weights $\theta^s$, and the second product operator consequently defines the distribution approximated by the client model $\epsilon_\theta^c$.

\begin{algorithm}[t]
    \begin{algorithmic}[1]
        \label{alg:training}
        \caption{Collaborative Training}
        \REQUIRE training dataset $D$; batch size $b$; number of time steps $T$, cut point $t_\zeta$,
        \\ variance scheduler $\alpha$, noise scheduler $\sigma$, clients $C$, server $s$; rounds $R$, local epochs $E$\\
    \STATE client dataset $D_c \subseteq D$
        \FOR{round $r = 1$ \TO $R$}
            \FOR{$c\in C$}
                \FOR{local epoch $e = 1$ \TO $E$}
                \FOR{Each batch $\{(x_0,y)\}^b \subseteq D_c$}
                    \STATE *** \textit{CLIENT NODE} ***
                    \STATE $t^c \sim \mathcal{U}[1,t_{\zeta}]^b$ and $t^s \sim \mathcal{U}[t_{\zeta},T]^b$
                    \STATE $\epsilon^c \sim \mathcal{N}(0, \mathbf{I})$, $\epsilon^s \sim \mathcal{N}(0, \mathbf{I})$
                    \STATE $x_{t^c}      \gets \alpha(t^c) \cdot x_0 + \sigma(t^c) \cdot \epsilon^c$\tikzmark{top}
                    \STATE $x_{t_{\zeta}} \gets \alpha(t_{\zeta}) \cdot x_0 + \sigma(t_{\zeta}) \cdot \epsilon^c$
                    \STATE $x_{t^s} \gets \alpha(t^s) \cdot x_{t_{\zeta}} + \sigma(t^s) \cdot \epsilon^s$ \tikzmark{middle1}
                    
                    \STATE $\mathcal{L}_{t^c} = || \epsilon_{\theta^c}(x_{t^c}, {t^c}, y) - \epsilon^c ||^2_2$\tikzmark{middle2}    \tikzmark{right}
                    \STATE Update $\theta^c$
                    \STATE Pass $x_{t^s}$ and $\epsilon^s$ to server $s$
                   \STATE                    *** \textit{SERVER NODE} ***
                    \STATE $\mathcal{L}_{t^s} = || \epsilon_{\theta^s}(x_{t^s}, {t^s}, y) - \epsilon^s ||^2_2$
                    \STATE Update $\theta^s$\tikzmark{bottom}
                \ENDFOR
                \ENDFOR
            \ENDFOR
        \ENDFOR
    \end{algorithmic}
    \AddNote{top}{middle1}{right}{\hspace{-10cm}Diffusion \\ \hspace{0cm}Process}
    \AddNote{middle2}{bottom}{right}{\hspace{-10cm}Denoising \\ \hspace{0cm}Process}
    \vspace{-0.4cm}
\end{algorithm}

\subsection{Collaborative Training} 
During training, client models $\epsilon^c_\theta$ and the server model $\epsilon^s_\theta$ are updated independently.
Alg.~1 provides our collaborative training procedure in pseudocode, which we now describe in more detail.
For training, each client $c\in C$ has access to a private dataset $D_c=\{(x^i_0,y^i)\}$ of images $x^i_0$ with optional textual attribute labels $y^i$.
In principle, our approach also works with unlabeled data and other kinds of labels, e.g., one-hot encoded label vectors and segmentation maps.
However, we focus on the use case of attribute-conditioned image generation in this work.
By using textual feature descriptions as labels, our implementation can easily be extended to more elaborate text-guided image synthesis.
As for the standard diffusion training process, each client samples a training batch $\{(x_0,y)\}^b \subseteq D_c$ of batch size $b$ together with client time steps $t^c \sim \mathcal{U}[1, t_\zeta]^b$ during each training step.
Gaussian noise is added to each training sample based on $t^c$ following the diffusion process defined in \cref{eq:diffusion}.
Note that $x_{t^c}$ and the cut-point image $x_{t_\zeta}$ are diffused from the same noise draw $\epsilon^c$ (Alg.~1, lines~9--10), so that the client trajectory and the sample handed to the server share a consistent noise realisation.
All noisy images are fed into the client's model to predict the added noise and update the model's parameters according to the loss function defined in \cref{eq:loss_fkt}.
In addition, each client uses the diffused image $x_{t_\zeta}$ from the cut point and samples additional server time steps $t^s \sim \mathcal{U}[t_\zeta, T]^b$ to provide the noisy images $x_{t^s}$ for the server.
The final noise image and the noise added to $x_{t_\zeta}$ are then used to update the server model's weights analogously.
We note that the process of adding additional noise for the server could, in principle, also be performed on the server side.
Line~11 constructs the server's training input by treating the cut-point sample $x_{t_\zeta}$ as a starting point and re-diffusing it to level $t^s$ with the standard marginal coefficients $\alpha(t^s),\sigma(t^s)$, rather than applying the exact conditional transition $q(x_{t^s}\mid x_{t_\zeta})$.
This is a deliberate design choice rather than an approximation error: the server is trained on exactly the marginal it is later asked to reverse, so the construction is internally consistent. Because the server's reverse trajectory at inference is initialised from pure noise and run over $[t_\zeta, T]$ on the same schedule (Alg.~2), the training and sampling marginals coincide.
We use a single shared noise schedule throughout; the only distinction between client and server is the timestep interval each is responsible for.

Regarding privacy, the client passes both $x_{t^s}$ and $\epsilon^s$, so the server can algebraically recover the cut-point representation $x_{t_\zeta}$.
However, under our threat model it does not observe the clean image $x_0$.
We therefore interpret the privacy benefit as follows: the server is limited to the noisy cut-point representation $x_{t_\zeta}$, and we evaluate the residual leakage from that representation empirically in \cref{sec:results}.

\begin{algorithm}[t]
    \begin{algorithmic}[1]
        \label{alg:inference}
        \caption{Collaborative Inference}
        \REQUIRE number of time steps $T$, label $y$, cut point $t_\zeta$, client $c$, server $s$
        \STATE Sample initial noise: $x_T \sim \mathcal{N}(0, \mathbf{I})$
        \STATE Choose client start $M \in [t_\zeta, T]$, e.g.\ $M = \left\lfloor t_\zeta + \tfrac{t_\zeta}{T}(T-t_\zeta)\right\rfloor$
        \STATE $t^c_{list} = \text{linearly spaced array generator}(1, M, t_\zeta)$
        \STATE $t \gets T$
        \WHILE{$t \geq 1$}
            \IF{$t > t_\zeta$}
                \STATE Compute $x_{t-1}$ using $\epsilon_{\theta^s}(x_t, t, y)$ and $\alpha(t), \sigma(t)$ \COMMENT{server: $T-t_\zeta$ steps}
            \ELSE
                \STATE Compute $x_{t-1}$ using $\epsilon_{\theta^c}(x_t, t^c_{list}[\,t\,], y)$ and $\alpha(t^c_{list}[\,t\,]), \sigma(t^c_{list}[\,t\,])$ \COMMENT{client: $t_\zeta$ steps; $t^c_{list}$ has $t_\zeta$ entries spanning $[1,M]$, $1$-indexed, $t\in\{1,\dots,t_\zeta\}$}
            \ENDIF
        \STATE $t \leftarrow t - 1$
        \ENDWHILE
    \end{algorithmic}
\end{algorithm}

\subsection{Collaborative Inference} 
After training, each client $c\in C$ can send a request to the server containing optional textual attribute labels $y$.
The individual steps during the inference are specified in Alg.~2.
The denoising loop runs $t$ from $T$ down to $1$: the server performs the first $T-t_\zeta$ steps using $\epsilon_\theta^s$ conditioned on the label $y$, after which the still noisy samples $\hat{x}_{t_\zeta}^s$ are handed to the client $c$, which performs the remaining $t_\zeta$ steps using its local model $\epsilon_\theta^c$.
To account for the increased amount of noise in $\hat{x}_{t_\zeta}^s$ and hence allow for a higher noise reduction on the client node, the variance and noise scheduler are adapted on the client side.
The client schedule need not start exactly at $t_\zeta$: it begins at a chosen value $M \in [t_\zeta, T]$ and is linearly spaced over its $t_\zeta$ steps, so the client compensates for the elevated residual noise without changing the total number of timesteps.
In our experiments, we set $M = \lfloor t_\zeta + \tfrac{t_\zeta}{T}(T-t_\zeta)\rfloor$, which extends the client interval slightly above $t_\zeta$; the simpler choice $M = t_\zeta$ (no rescaling) is equally valid and recovers the unadjusted schedule.
A sufficiently large cut point $t_\zeta$ ensures that possibly sensitive features are generated on the client side while the server performs the less privacy-critical initial denoising steps.
If multiple clients request samples from the same label~$y$, the server-side denoising process can be run once to generate an intermediate noise sample, and each client solely has to compute the remaining denoising steps.

\section{Experiment}\label{sec:experiments}
To assess our approach, we implement Alg.~1 and Alg.~2 and train the models on common benchmark datasets. We simulate a scenario in which $k=5$ clients use a trusted server to train a DDPM or LDM collaboratively. Each client has access to an individual subset from the same domain. Thereby, we investigate the influence of collaborative training on the fidelity of generated images. Furthermore, we analyze the influence of the chosen cut point $t_\zeta$ on sample fidelity with respect to disclosed information.
\begin{figure}
    \centering
    \includegraphics[width=
    \linewidth]{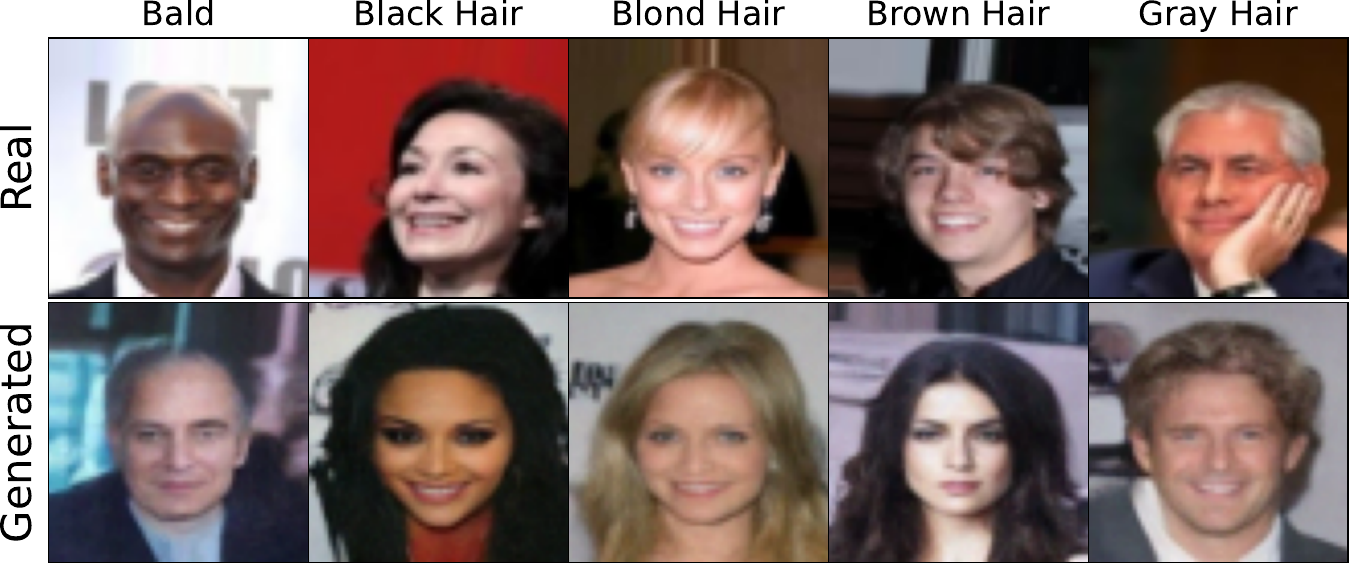}
    \caption{Comparison between random samples from the training set (top row) and images generated with our collaborative diffusion models trained with cut point $t_\zeta=100$ (bottom row). Images were not cherry-picked and generated starting with the same initial noise. The results demonstrate that collaboratively training diffusion models can achieve high image quality and attribute fidelity.}
    \label{fig:attribute_illustrations}
    \Description{A two-row grid of CelebA face images with five columns labeled Bald, Black Hair, Blond Hair, Brown Hair, and Gray Hair.
    The top row, labeled ``Real,'' shows actual training-set photographs of faces with each hair attribute.
    The bottom row, labeled ``Generated,'' shows faces synthesized by CollaFuse at cut point t-zeta equals 100, each matching the corresponding column attribute.
    Faces in both rows were generated from the same initial noise and were not cherry-picked, illustrating that the collaborative model reproduces the target hair-color attributes with realistic quality.}
\end{figure}
\subsection{Experimental Protocol}
To ensure consistent conditions and avoid confounding factors, we maintain identical training and inference hyperparameters and seeds across different runs and settings, if not stated otherwise.
We provide our source code for reproducibility using a GitHub repository\footnote{\url{https://github.com/SimeonAllmendinger/collafuse}}.
\begin{figure}[t]
    \centering
    \includegraphics[width=\linewidth]{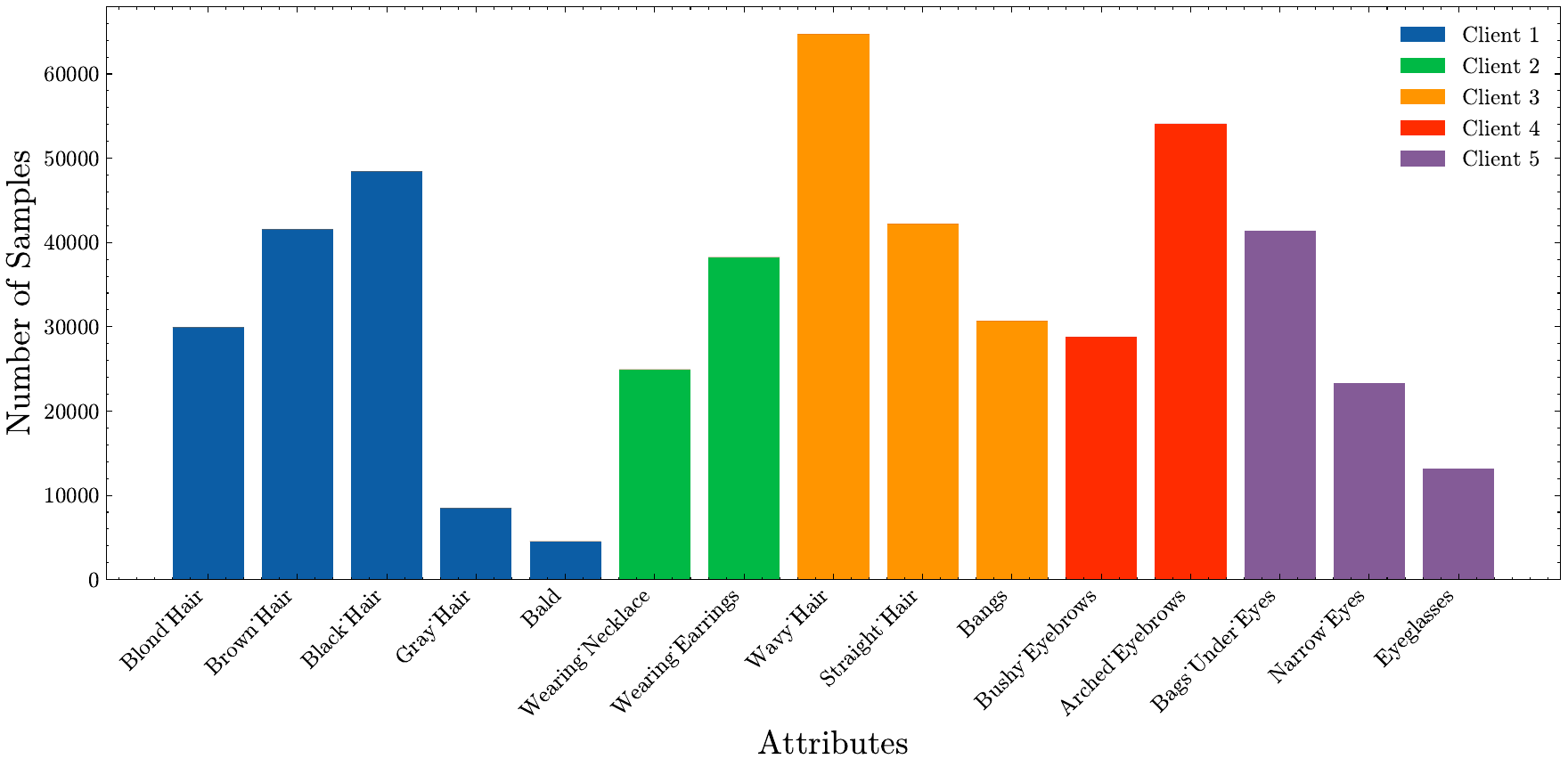}
    \caption{While datasets such as CIFAR-10 and AWA2 were evenly partitioned among clients (IID), CelebA was deliberately distributed according to attribute labels to simulate heterogeneous, non-IID client data.
    Each color indicates one client's subset of dominant attributes, such as hair color, facial features, or accessories.
    This setup mimics realistic data heterogeneity across participants, where clients capture distinct subpopulations rather than uniform samples.}
    \label{fig:cleba_data_distribution}
    \Description{A vertical bar chart. The x-axis lists fifteen CelebA attribute names (such as Blond Hair, Brown Hair, Black Hair, Gray Hair, Bald, Wearing Necklace, Wearing Earrings, Wavy Hair, Straight Hair, Bangs, Bushy Eyebrows, Arched Eyebrows, Bags Under Eyes, Narrow Eyes, Eyeglasses). The y-axis is number of samples, from 0 to about 64,000. Each bar is colored by one of five clients. The chart shows a deliberately non-IID partition: each client's color dominates a distinct subset of attributes rather than appearing uniformly, so clients hold heterogeneous attribute distributions. Bar heights vary widely, with Wavy Hair the tallest at roughly 64,000.}
\end{figure}


\noindent\textbf{Model Architecture:} We adopt the network architectures of traditional DDPMs~\citep{Ho2020}, Imagen Models~\citep{saharia2022photorealistic}, and LDMs~\citep{Rombach_2022_CVPR} which are based on the U-Net architecture~\citep{ronneberger15unet} to process $32\times 32$,  $64\times 64$, $256\times 256$, and $512\times 512$ RGB images. As in the original implementation, the Imagen models are conditioned on text embeddings computed by T5-Base~\citep{2020t5}.
For the DDPMs, we use one-hot encoding.
For the LDMs, we use pre-trained U-Nets as well as text- and autoencoders from ``\textit{CompVis/ldm-text2im-large-256}''~\citep{Rombach_2022_CVPR} and ``\textit{runwayml/stable-diffusion-v1-5}''~\citep{Rombach_2022_CVPR} from huggingface. Unlike~\citet{saharia2022photorealistic}, we do not apply any super-resolution model to increase the fidelity of the generated images in the Imagen model, as the focus of our work lies on the feasibility of the collaborative training and inference process.
However, our collaborative diffusion setting also allows each client to add individual or shared super-resolution models~\citep{chitwan22palette} to their pipeline to upscale the generated images.

\noindent\textbf{Datasets:} We train and evaluate our collaborative diffusion models on common datasets: \textit{CelebA}, \textit{CIFAR-10}, \textit{Animals-with-Attributes2 (Awa2)}.
The CelebA dataset, collected by~\citet{celeba}, consists of over 200,000 facial images of celebrities annotated with 40 binary attribute labels, including gender appearance, perceived age, hair color, and facial expressions.
The CIFAR-10 dataset~\citep{cifar10} includes 60,000 32x32 color images across 10 classes, such as airplanes, automobiles, and animals.
The Awa2 dataset~\citep{Awa2} contains 37,322 images of 50 animal classes with 85 attributes.

To evaluate \texttt{CollaFuse} under more realistic conditions, we extended our experimental setup beyond IID data partitions.
While CIFAR-10 and Awa2 were uniformly split among clients, the CelebA dataset was distributed non-IID with respect to semantic attributes (\Cref{fig:cleba_data_distribution}).
Each client thus specialized in distinct attribute combinations (e.g., hair color, facial traits, or accessories), reflecting natural data heterogeneity often observed in federated or collaborative learning scenarios.

\noindent\textbf{Training Parameters:}
Our training protocol employs a learning rate of 0.001, a batch size of 8, and $T=1000$ timesteps.
Each client holds 10,000 / 56,870-123,908 / 6,717 training images and 2,000 / 6,319-13,768 / 746 test images (hold-out dataset) w.r.t the different datasets CIFAR-10 / CelebA / Awa2.
For CelebA, the attribute-based non-IID split is overlapping rather than a disjoint partition: because each image carries multiple binary attributes, an image can be assigned to more than one client's attribute band, so the per-client counts sum to more than the dataset size.
The experiments were conducted utilizing 11 NVIDIA A100-SXM4-80GB for computational processing.

\begin{figure*}[t]
    \centering
    \includegraphics[width=\linewidth]{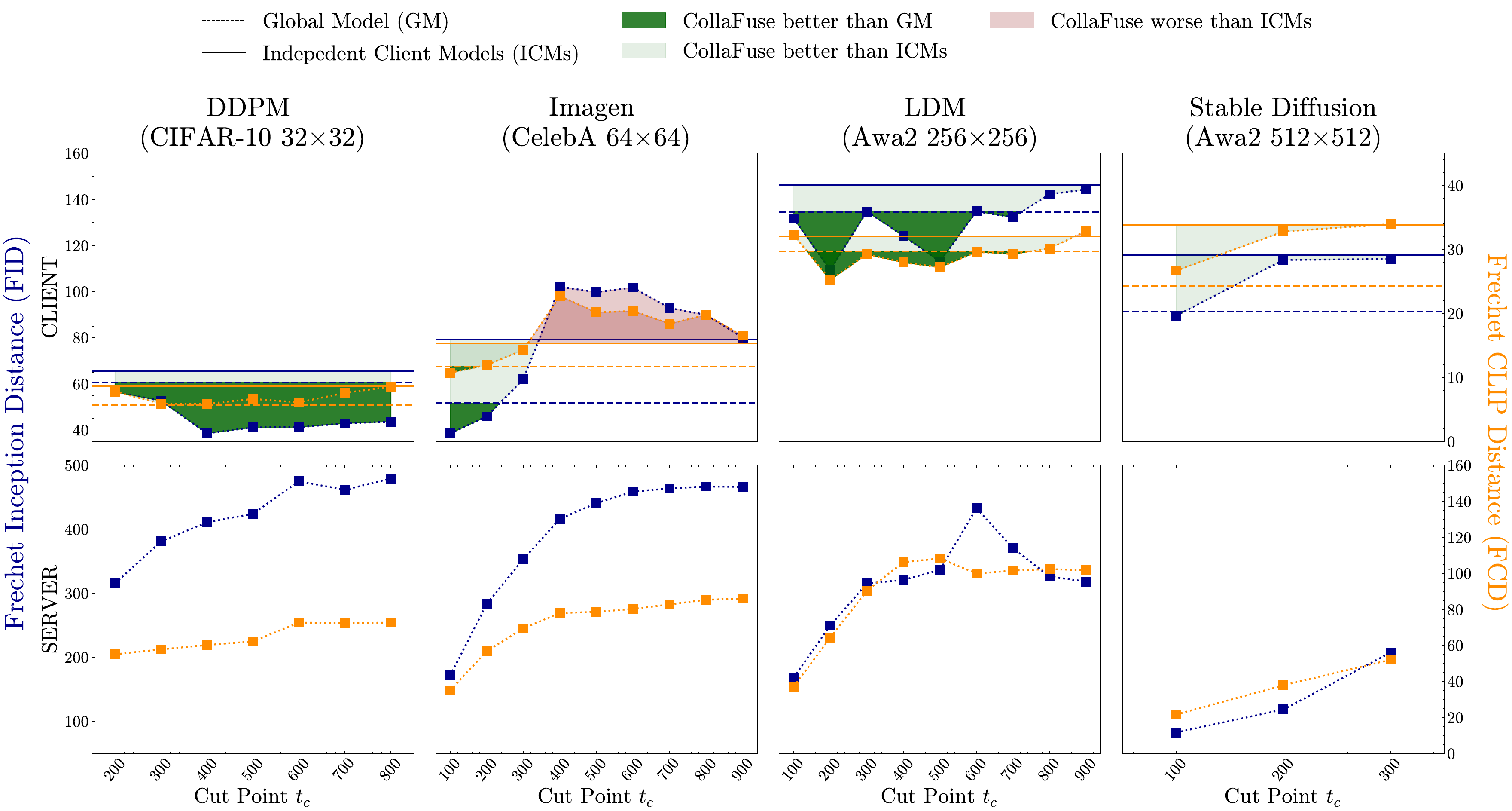}
    \caption{Fidelity evaluation of clients and server using FID$\downarrow$ and FCD$\downarrow$: We assess 10,000 / 6,319-13,768 / 3,730 real $x_0$ and collaboratively generated images ($\hat{x}_0^c \circ \hat{x}^s_{t_\zeta}(\epsilon)$) from the CIFAR-10 / CelebA / Awa2 dataset across clients. Our findings show that cut points with $t_\zeta \leq 200$ outperform the baseline of independent client models (ICMs) ($t_\zeta = 1000$).
    In some of these settings---particularly at lower resolutions or smaller latent sizes---small cut points also surpass the global model (GM) ($t_\zeta = 0$), which is trained on the pooled data of all clients and is therefore not a privacy-feasible baseline in our setting.
    As expected, information disclosure almost consistently decreases as the cut point ($t_\zeta$) increases across all models (DDPM, Imagen, LDM, Stable Diffusion).}
    \label{fig:performance_clients}
    \Description{A four-column by two-row grid of line/scatter plots comparing image fidelity across architectures.
    Columns are DDPM on CIFAR-10 ($32\times32$), Imagen on CelebA (64×64), LDM on Awa2 ($256\times256$), and Stable Diffusion on Awa2 ($512\times512$).
    The top row shows client-side Fréchet Inception Distance and Fréchet CLIP Distance versus cut point; the bottom row shows the same metrics for server-side outputs.
    Square markers give scores at each cut point. Horizontal lines mark two baselines: a dashed line for the pooled global model and a solid line for independent client models.
    Shaded regions indicate where CollaFuse beats the global model, beats independent client models, or does worse than independent client models.
    Across plots, cut points at or below 200 generally outperform independent client models, while server-side distances rise as the cut point increases.}
\end{figure*}

\noindent\textbf{Evaluation Metrics:} To assess the quality of the generated images, we calculate the common \textit{Fréchet Inception Distance} (FID)~\citep{NIPS2017_8a1d6947} and the \textit{Fréchet CLIP Distance} (FCD) between the test dataset and generated images from each client. We differentiate between images generated by clients from pure Gaussian noise ($t_\zeta = 1000$), and images generated based on the server image at the cut point. All metrics are computed on the implementations provided by~\citet{parmar2021cleanfid} to ensure stable and comparable evaluation results. For both metrics, lower values indicate better approximation of the training distribution and improved image quality.

As we are interested in the performance of \texttt{CollaFuse}, we calculate the image fidelity across the set of clients to compare the performance for different values of $t_\zeta$.
Furthermore, we calculate the fidelity between the partially diffused images at the cut point $t_\zeta$ and the original images of the clients to assess the information disclosed to the server.
We use this as a \emph{proxy for the visual and distributional similarity} between the server-side intermediate outputs and real images, rather than as a privacy metric: a high FID/FCD indicates that server outputs are visually dissimilar from real data, but does not by itself bound membership, reconstruction, or attribute leakage.
We complement it with the dedicated attribute-inference and inversion experiments in \cref{sec:results}, which assess leakage directly.

\subsection{Experimental Results}\label{sec:results}
 \begin{figure}[t]
    \centering
    \includegraphics[width=\linewidth]{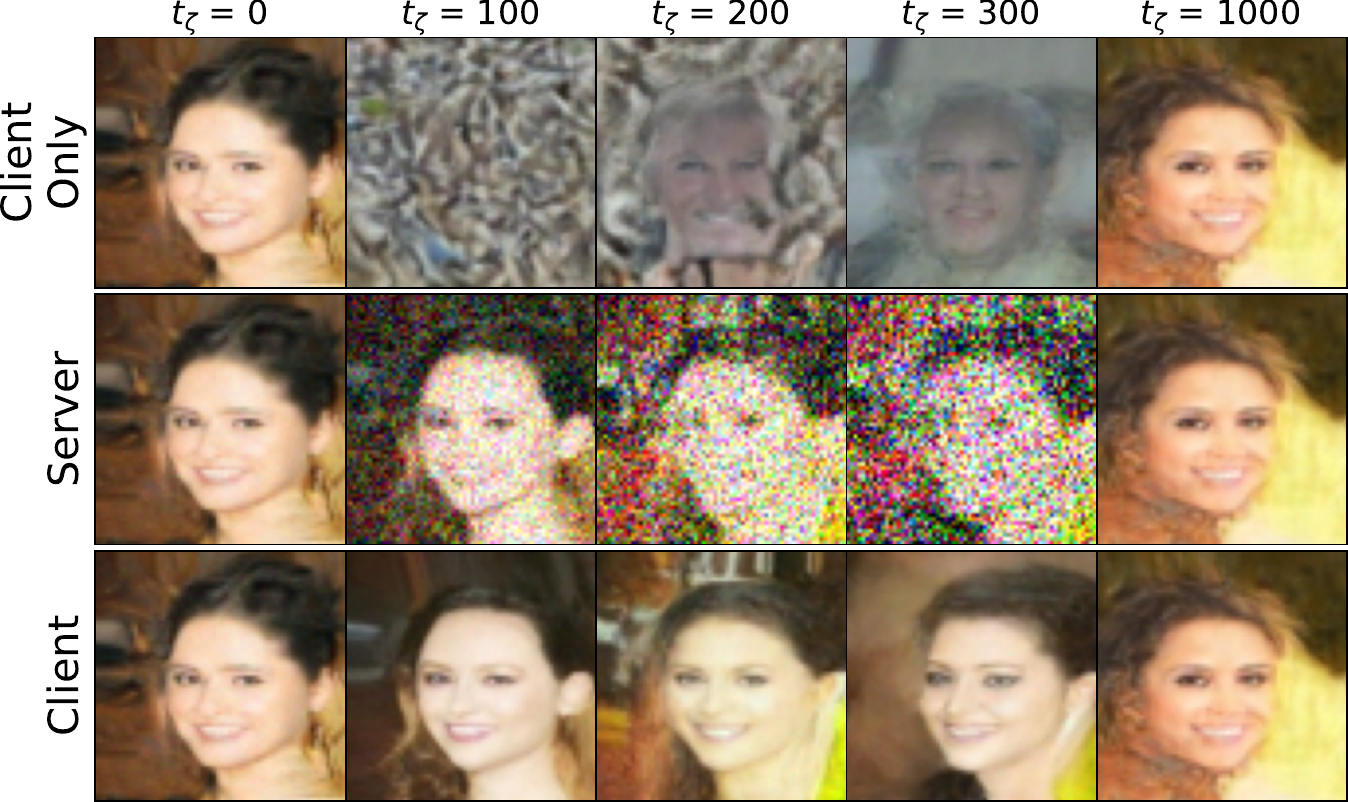}
    \caption{Samples generated by \texttt{CollaFuse} trained on CelebA with different cut points $t_\zeta$. The top row depicts images produced by the server, which are then sent to the client. The bottom row shows the samples after the final denoising performed by the client. For $t_\zeta=0$, the server performs the full denoising process; for $t_\zeta=1000$, each client trains a separate diffusion model without a server component.}
   \label{fig:sample_illustrations}
   \Description{
   A three-row by five-column grid of CelebA face images at cut points t-zeta of 0, 100, 200, 300, and 1000. The top row, ``Client Only,'' shows what a client produces alone.
   The middle row, ``Server,'' shows the partially denoised image the server hands to the client; it grows noisier and more degraded as the cut point increases, becoming near-random color noise at 200 and 300.
   The bottom row, ``Client,'' shows the final image after client denoising, which remains a recognizable face across cut points. At cut point 0 the server does everything; at 1000 each client works alone with no server.
   }
\end{figure}
In our experiments, we analyze the performance of collaborative diffusion models by focusing on specific distributed categories among clients. \Cref{fig:attribute_illustrations} displays exemplary generated images of our collaborative approach from the dataset CelebA, comparing them with their attributes and real images from the dataset. Our quantitative analysis includes a comparison with two baselines: global model ($t_\zeta = 0$) and independent client models ($t_\zeta = 1000$). The single global model (GM) on the server node is trained using the combined datasets of all clients, while the independent client models (ICM) are trained on client-specific distributed sub-datasets and separately operate on the client node.
The FID and FCD scores in \Cref{fig:performance_clients} illustrate the feasibility of training diffusion models collaboratively. Each column shows the scores for a different architecture, dataset, and image resolution. The first row compares the performance for different cut-points with the GM (dashed horizontal line) and ICMs (solid horizontal line).
The squares represent the FID and FCD scores for the respective cut points.
Further, it is illustrated whether collaboration increases or decreases the performance. The second row shows the FID and FCD scores based on the images generated from the server, which are then sent to the clients.
The figure shows that models with cut points $t_\zeta \leq 200$ surpass the performance of the independent client models in favor of collaborative image synthesis.
In some settings, our collaborative experiments with smaller cut points also outperform the global model, which pools client data and is therefore not a privacy-feasible baseline.
However, cut points $t_\zeta > 300$ often weaken the fidelity of image generation, which converges with the performance of the independent client models for higher cut points.

In terms of \textbf{image quality}, we observe that the degradation of image fidelity and visual characteristics is evident in images generated by the server as the cut point ${t_\zeta}$ increases. \Cref{fig:sample_illustrations} displays this deterioration, which affects the images generated by the server at higher cut points. Conversely, the client's performance in generating images from noise is compromised when operating with low cut points. However, leveraging our collaborative approach, each client can produce meaningful images with improved fidelity, as shown in~\Cref{fig:performance_clients}, making use of denoised images provided by the server.
Additionally, our results indicate that incorporating an adjustment of the variance and noise scheduler into the collaborative inference process significantly enhances the denoising capabilities on the client node.
This is particularly effective given the higher levels of residual noise in the images received from the server, leading to an improvement in the overall quality of the images. 
Adopting the adjusted parameter $M$ in Alg.~2, especially with an increased emphasis on managing variance and noise in the collaborative setting, proves to be beneficial in refining the quality of the final image output.
To put these results into context, we perform ablation studies that compare \texttt{CollaFuse} to FedAvg-DDPM in terms of image quality, communication cost, and client-side computational cost during training and inference, showing that \texttt{CollaFuse} consistently improves KID across four datasets while reducing client-side FLOPs to $t_\zeta/T$ of the FedAvg-DDPM cost, with details provided in~\Cref{app:cost-analysis}.

\begin{figure}[t]
    \centering
    \includegraphics[width=\linewidth]{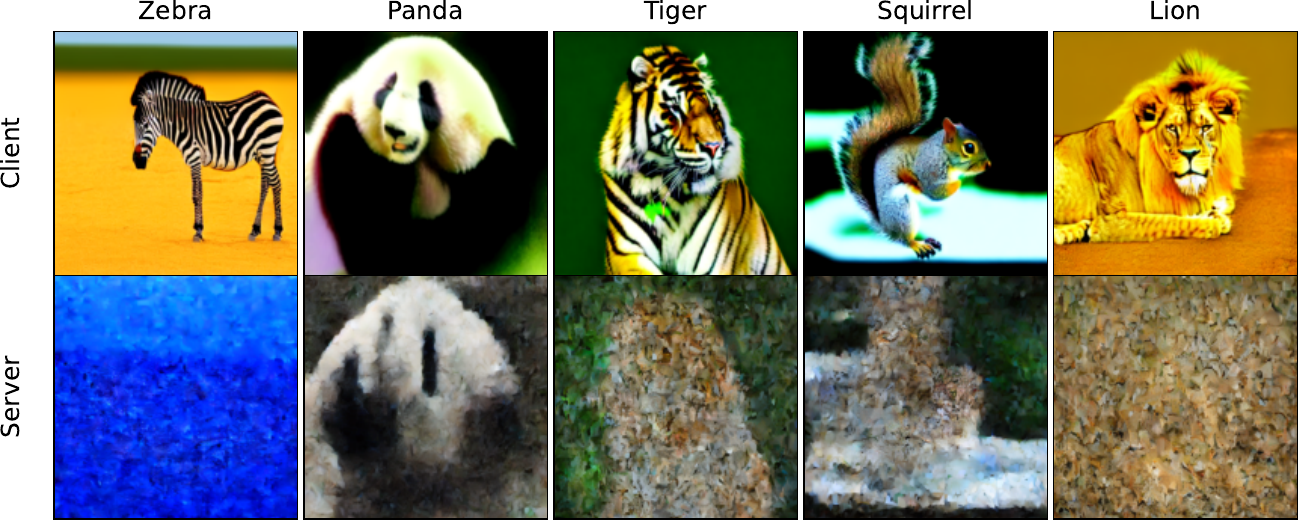}
    \caption{Samples generated by \texttt{CollaFuse} trained on Awa2 for the cut point $t_\zeta = 400$. The top row depicts images after the final denoising performed by the client. The bottom row shows images produced by the server, which are then sent to the client.}
   \label{fig:sample_illustrations_Awa2}
   \Description{A two-row by five-column grid of Awa2 animal images at cut point t-zeta equals 400, with columns Zebra, Panda, Tiger, Squirrel, and Lion.
   The top row, ``Client,'' shows the final high-quality animal images after client denoising.
   The bottom row, ``Server,'' shows the corresponding server outputs handed to the client, which are blurry, low-detail textures with little recognizable animal structure, illustrating how little identifiable content the server discloses at this cut point.}
\end{figure}

In terms of \textbf{information disclosure} to the server, we analyze the similarity between real images and the images generated by the server--these are the images sent to clients for further denoising. We use the FID and FCD scores to assess how similar the generated images match the training distribution. The results show that as the cut point increases, the FID and FCD scores also rise, indicating a decrease in the similarity between the distribution of real images and that of the server-generated images. \Cref{fig:sample_illustrations_Awa2} illustrates the server's ability to generate images and the extent of information disclosed for a cut point \( t_\zeta = 400 \).

\begin{figure}[t]
    \centering
    \includegraphics[width=\linewidth]{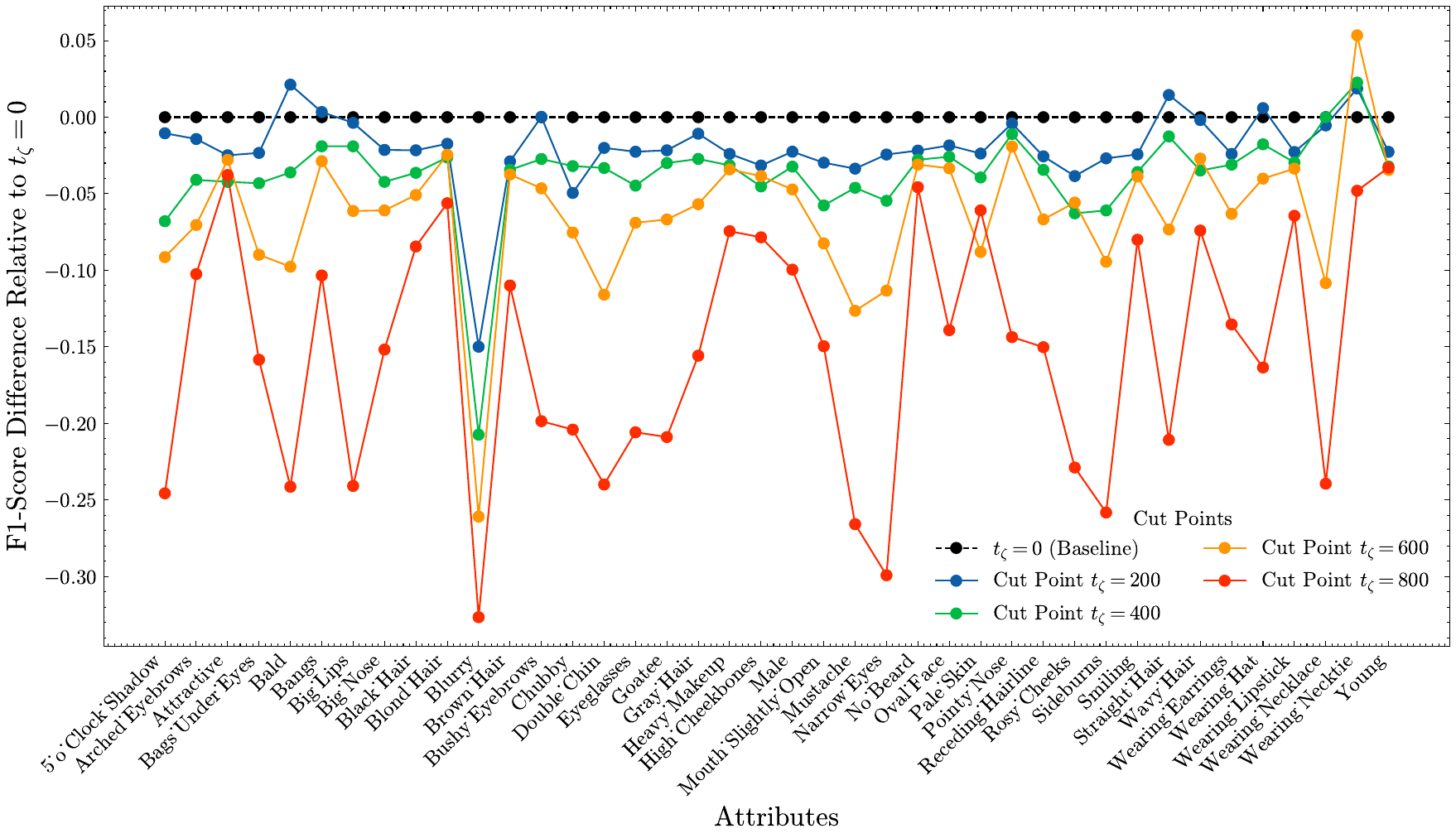}
    \caption{Attribute inference F1-score differences for 40 CelebA attributes across varying cut points $t_\zeta$.
    Each line represents the change in F1-score relative to the baseline without a cut point ($t_\zeta=0$).
    Lower scores indicate reduced attribute predictability from intermediate representations.
    The results show a consistent decline in inference accuracy for earlier cut points, suggesting that higher noise levels at intermediate stages naturally mitigate information leakage. All baseline F1-scores at $t_{\zeta} = 0$ are above 65\%.
    }
    \label{fig:attribute-inference}
    \Description{A line plot of attribute-inference F1-score difference (y-axis, ranging roughly from minus 0.30 to plus 0.05) across forty CelebA attribute names on the x-axis.
    Five lines correspond to cut points 0 (baseline, flat at zero), 200, 400, 600, and 800.
    Lines for larger cut points sit progressively lower, showing that attribute predictability from intermediate server representations declines as the cut point increases.
    The 800 line drops most, reaching about minus 0.30 for some attributes. All baseline F1-scores at cut point 0 exceed 65 percent.}
\end{figure}

We assume an honest-but-curious setting: each party follows the protocol but may inspect what it receives.
The server sees $x_{t^s}$ and $\epsilon^s$ per sample (Alg.~1, line~14), from which it can recover $x_{t_\zeta}$ but not $x_0$; a malicious client sees only the intermediate sample $\hat{x}^s_{t_\zeta}$ for its own label and its own data.
Our attribute-inference adversary is a (non-pretrained) ViT-Base classifier trained on intermediate images; our inversion adversary is a simulated client reconstructing another client's data from cut-point features.
We do not consider colluding, adaptive, or white-box adversaries, which we leave to future work.
To further substantiate our analysis of information disclosure at intermediate cut points, we conducted a dedicated attribute inference experiment on CelebA.
Using a ViT-Base classifier (\textit{google/vit-base-patch16-224}; not pre-trained), we trained attribute predictors on intermediate images generated at different cut points $t_\zeta$.
\Cref{fig:attribute-inference} summarizes the resulting F1-score differences for all 40 attributes relative to the baseline without a cut point ($t_\zeta=0$).
The results reveal a clear decline in attribute inference accuracy as the cut point increases, confirming that as more steps are assigned to the client the server-side representation is taken earlier in the reverse process, contains more noise, and leaks substantially less semantic and identity-related information.
Under the specific classifiers and attacks tested, earlier (noisier) representations show reduced leakage.
These are attack-based empirical observations under the stated threat model, not formal guarantees, and stronger adversaries may extract more.

Building on the attribute inference analysis, which focused on the leakage of semantic or sensitive attribute information, we next investigated whether intermediate representations could be exploited to reconstruct original data.
This complementary evaluation assesses a more direct form of privacy risk, namely, the ability of an adversarial or malicious client to recover identifiable visual content from shared features.

\begin{figure}[t]
    \centering
    \includegraphics[width=1.0\linewidth]{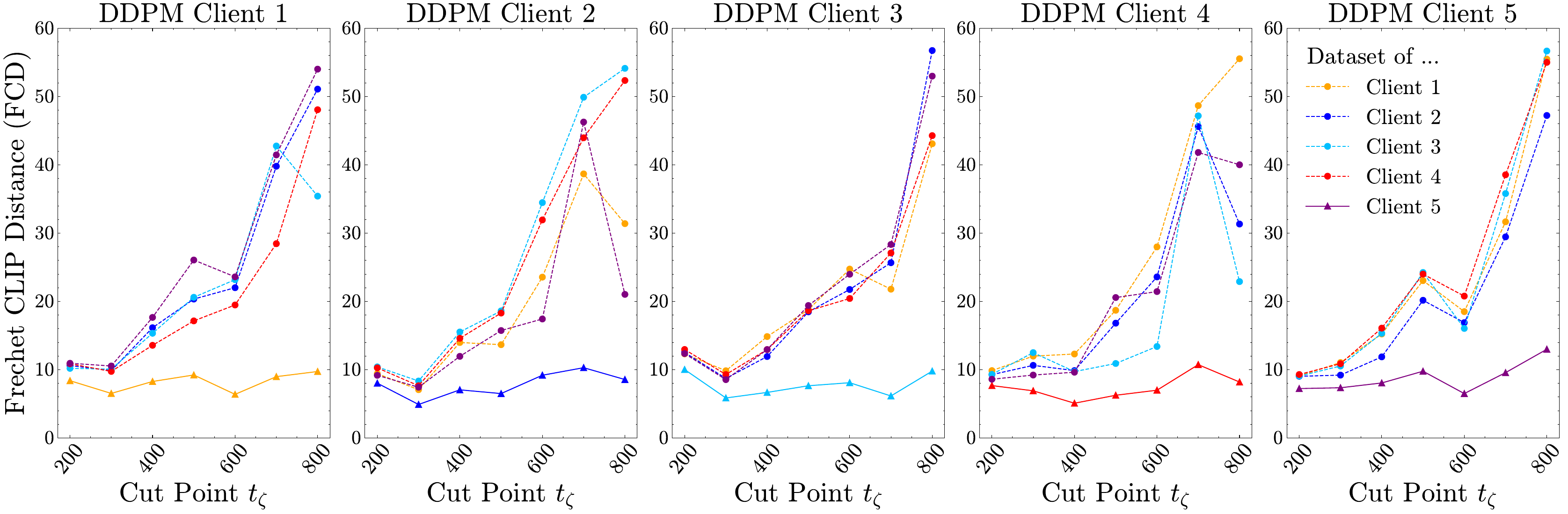}
    \caption{Results of cross-client inversion attacks on CIFAR-10 across different cut points $t_\zeta$.
    As the cut point increases (more denoising steps are performed on the client and the server hands over an earlier, noisier sample), reconstructed images exhibit a marked loss of client-specific semantics and visual detail.
    For $t_\zeta \geq 400$, models show a decline in their ability to reconstruct other clients' training data, reflected in an increased FCD metric.}
    \label{fig:adverserial_clients}
    \Description{A row of five line plots, one per DDPM client (Client 1 through Client 5), showing cross-client inversion-attack results on CIFAR-10.
    Each plot's x-axis is the cut point (200 to 800) and y-axis is Fréchet CLIP Distance (0 to 60). 
    Within each plot, five colored lines represent attempts to reconstruct each of the five clients' datasets.
    Reconstruction of a model's own data stays low (good reconstruction) while reconstruction of other clients' data rises sharply as the cut point increases, especially beyond 400, indicating reduced cross-client information leakage at higher cut points.}
\end{figure}

To further assess reconstructibility and potential cross-client information leakage, we performed inversion attack experiments on the CIFAR-10 dataset.
In this setting, a simulated malicious client attempted to reconstruct another client's private (training) data or infer its images from intermediate representations shared during collaborative generation.
Using DDPMs conditioned on features extracted at various cut points $t_\zeta$, we observed that reconstructed samples rapidly lost identifiable content as the cut point moved earlier in the generation process (\Cref{fig:adverserial_clients}).
For cut points $t_\zeta \geq 400$, the models exhibited a pronounced decline in their ability to reconstruct training data from other clients.
Specifically, reconstruction quality remained higher for a model's own training data but deteriorated substantially for data originating from different clients, resulting in a sharp increase in the FCD metric.
This indicates that, beyond this cut point, adversarial clients were increasingly constrained in reproducing the data distribution of other participants, reflecting a reduced potential for cross-client information leakage.
Under the inversion attack evaluated here, these findings indicate that \texttt{CollaFuse} constrains cross-client reconstruction at higher cut points, consistent with the attribute-inference results. As with the attribute-inference analysis, this is empirical evidence against the specific attacker defined in our threat model rather than a formal guarantee.

\section{Discussion, Limitations, and Future Work}\label{sec:discussion}

Our findings reveal that \texttt{CollaFuse} produces images of higher quality than those generated by independently trained diffusion models using sub-datasets.
In several settings it also matches or surpasses globally trained
diffusion models, which pool client data and are therefore not a
privacy-feasible baseline. 
This highlights the efficacy of our method and demonstrates its capability to tackle the personalization versus generalization challenge frequently encountered in FL scenarios~\citep{Yao2024}.
Beyond image fidelity, our extended experiments provide a more comprehensive understanding of \texttt{CollaFuse} in terms of information disclosure.
Under the attacks we evaluate, the attribute-inference analysis on CelebA shows that intermediate representations at higher cut points reveal substantially less semantic and identity-related information, consistent with higher noise levels in the diffusion process acting as a privacy buffer. 
Complementary inversion attack experiments on CIFAR-10 show the same trend: as the cut point increases, reconstructed samples rapidly lose visual fidelity and class-specific detail.
Taken together, these results provide \emph{attack-based empirical evidence} that, against the specific adversaries tested, the collaborative exchange in \texttt{CollaFuse} discloses limited information and constrains cross-client reconstruction.
We emphasise that these are empirical observations under a defined threat model rather than formal privacy guarantees; stronger or adaptive adversaries may extract more, and we leave provable bounds to future
work (see below).
At the same time, our approach circumvents the need to share raw data or complete model updates, opting instead to share only the diffused images alongside server-side noise.
Furthermore, by delegating computationally intensive tasks, our approach significantly reduces the computational load on individual clients, leveraging the potential strengths of growing foundation models on large servers (see~\Cref{app:cost-analysis}).
Consequently, even where image fidelity decreases, clients may still favor a collaborative diffusion model for its ability to lighten their computational load, even in simple one-to-one setups.
Thus, our strategy facilitates a balanced optimization of performance, information disclosure, and computational resources and can be tailored to individual preferences.

\noindent\textbf{Limitations:}
Our current implementation of \texttt{CollaFuse} is based on DDPMs, which provide a theoretically grounded framework for analyzing collaborative training dynamics, cut-point sensitivity, and information disclosure. 
However, DDPMs are computationally intensive due to their stochastic denoising process.
While this choice enables interpretability and a rigorous analysis of privacy aspects, it limits inference efficiency. 
Furthermore, our current approach assumes trustworthy clients and an honest server. However, collaborative frameworks are known to be susceptible to backdoor attacks~\citep{wang20attacktails,tajalli23serversidebackdoor}, in which an adversarial client may attempt to inject hidden functionalities into a shared model.
Prior research has shown that diffusion models themselves are vulnerable to such attacks~\citep{struppek2022rickrolling,chou23diffusionbackdoor}.
While these scenarios are beyond the current scope, future work will extend our framework to systematically assess and mitigate adversarial threats while preserving image fidelity and efficiency.

\noindent\textbf{Future Work:}
Future research will focus on extending \texttt{CollaFuse} to more advanced and efficient denoising paradigms such as DDIM~\citep{song2020denoising} and flow-based diffusion models.
These variants can significantly accelerate inference and reduce resource requirements while maintaining the collaborative and privacy-preserving characteristics of our approach.
In addition, we plan to scale \texttt{CollaFuse} to larger datasets, dynamic cut-point adaptation, and high-resolution domains to comprehensively assess its scalability and privacy–utility trade-offs under realistic conditions.
Building on our privacy evaluations, future work will further explore formal privacy guarantees and attack-based analyses to establish stronger theoretical and empirical bounds on information leakage, while also analyzing the trade-offs among efficiency, privacy, and generation quality in more detail---also against potential benchmarks like FedAvg-DDPMs.
Another promising direction is to investigate vulnerabilities such as backdoor attacks in collaborative diffusion and to develop countermeasures that preserve fidelity and efficiency.
Additionally, integrating differentially private training algorithms~\citep{dockhorn2022differentially,ghalebikesabi2023differentially} and studying memorization effects in collaborative diffusion~\citep{carlini23diffusion,burg21memorization,hintersdorf2024finding} will further advance \texttt{CollaFuse} toward scalable, secure, and privacy-aware real-world applications. Finally, future research should place greater emphasis on algorithmic fairness to reduce the reinforcement of stereotypes and cultural biases when implementing CollaFuse~\citep{Struppek2023}, especially since privacy and fairness are both essential characteristics of real-world collaborative learning~\citep{balbierer2024multivocal}.

\section{Conclusion}\label{sec:conclusion}

Our collaborative diffusion approach \texttt{CollaFuse} offers a novel solution to the challenges of diffusion-based generative models. By dividing the denoising process between a shared server and client models, we address performance, information disclosure, and computational concerns effectively. Our approach enables clients to outsource computationally intensive denoising steps to the server, balancing image quality without the necessity of sharing raw data. Through experiments, we demonstrate the effectiveness of collaborative training in enhancing image quality tailored to each client's domain while reducing the number of denoising steps on the client side. These findings highlight the potential of \texttt{CollaFuse} in advancing distributed machine learning research and development. 

\printbibliography

@InProceedings{McMahan2017FL,
  title = 	 {Communication-Efficient Learning of Deep Networks from Decentralized Data},
  author = 	 {McMahan, Brendan and Moore, Eider and Ramage, Daniel and Hampson, Seth and Arcas, Blaise Aguera y},
  booktitle = 	 {International Conference on Artificial Intelligence and Statistics (AISTATS)},
  pages = 	 {1273--1282},
  year = 	 {2017},
}

@article{cifar10,
    title= {CIFAR-10 (Canadian Institute for Advanced Research)},
    journal= {},
    author= {Alex Krizhevsky and Vinod Nair and Geoffrey Hinton},
    year= {2009},
    url= {http://www.cs.toronto.edu/~kriz/cifar.html},
    terms= {}
}

@ARTICLE{Awa2,
  author={Xian, Yongqin and Lampert, Christoph H. and Schiele, Bernt and Akata, Zeynep},
  journal={IEEE Transactions on Pattern Analysis and Machine Intelligence}, 
  title={Zero-Shot Learning—A Comprehensive Evaluation of the Good, the Bad and the Ugly}, 
  year={2019},
  volume={41},
  number={9},
  pages={2251-2265},
}

@inproceedings{shokri17meminf,
  author       = {Reza Shokri and
                  Marco Stronati and
                  Congzheng Song and
                  Vitaly Shmatikov},
  title        = {Membership Inference Attacks Against Machine Learning Models},
  booktitle    = {Symposium on Security and Privacy (S\&P)},
  pages        = {3--18},
  year         = {2017},
}

@inproceedings{zhu19leakage,
  author       = {Ligeng Zhu and
                  Zhijian Liu and
                  Song Han},
  title        = {Deep Leakage from Gradients},
  booktitle    = {Advances in Neural Information Processing Systems (NeurIPS)},
  pages        = {14747--14756},
  year         = {2019},
}

@article{Shen2023,
  title = {A Federated Learning System for Histopathology Image Analysis With an Orchestral Stain-Normalization GAN},
  volume = {42},
  number = {7},
  journal = {IEEE Transactions on Medical Imaging},
  author = {Shen,  Yiqing and Sowmya,  Arcot and Luo,  Yulin and Liang,  Xiaoyao and Shen,  Dinggang and Ke,  Jing},
  year = {2023},
  pages = {1969–1981}
}

@article{Nazir2023,
  title = {Federated Learning for Medical Image Analysis with Deep Neural Networks},
  volume = {13},
  number = {9},
  journal = {Diagnostics},
  author = {Nazir,  Sajid and Kaleem,  Mohammad},
  year = {2023},
}

@inproceedings{Schwermer2023FLEV,
author = {Schwermer, Ren\'{e} and Bicer, Ekin-Alp and Schirmer, Pascal and Mayer, Ruben and Jacobsen, Hans-Arno},
title = {Federated Computing in Electric Vehicles to Predict Coolant Temperature},
year = {2023},
booktitle = {International Middleware Conference: Industrial Track},
pages = {8–14},
}

@inproceedings{Schwerner2020FLEngergy,
author = {Schwermer, Ren\'{e} and Buchberger, Jonas and Mayer, Ruben and Jacobsen, Hans-Arno},
title = {Federated office plug-load identification for building management systems},
year = {2022},
booktitle = {ACM International Conference on Future Energy Systems},
pages = {114–126},
}

@article{Madhura2022,
author = {Joshi, Madhura and Pal, Ankit and Sankarasubbu, Malaikannan},
title = {Federated Learning for Healthcare Domain - Pipeline, Applications and Challenges},
year = {2022},
volume = {3},
number = {4},
journal = {ACM Transactions on Computing for Healthcare},
}

@article{Little2023,
  title = {Federated learning for generating synthetic data: a scoping review},
  volume = {8},
  number = {1},
  journal = {International Journal of Population Data Science},
  author = {Little,  Claire and Elliot,  Mark and Allmendinger,  Richard},
  year = {2023},
}

@inproceedings{Veeraragavan2023SecuringFLGAN,
author = {Veeraragavan, Narasimha Raghavan and Nyg\r{a}rd, Jan Franz},
title = {Securing Federated GANs: Enabling Synthetic Data Generation for Health Registry Consortiums},
year = {2023},
booktitle = {International Conference on Availability, Reliability and Security (ARES)},
  pages = {89:1--89:9},
}

@inproceedings{Kortoi2022,
  author = {Korto\c{c}i,  Pranvera and Liang,  Yilei and Zhou,  Pengyuan and Lee,  Lik-Hang and Mehrabi,  Abbas and Hui,  Pan and Tarkoma,  Sasu and Crowcroft,  Jon},
  title = {Federated split GANs},
  booktitle = {ACM Workshop on Data Privacy and Federated Learning Technologies for Mobile Edge Network},
  year = {2022},
}

@article{ohta2023lambdasplit,
  author       = {Shoki Ohta and Takayuki Nishio},
  title        = {{$\Lambda$-Split: A Privacy-Preserving
                  Split Computing Framework for Cloud-Powered Generative AI}},
  journal      = {arXiv preprint},
  eprint       = {2310.14651},
  archivePrefix= {arXiv},
  year         = {2023},
}

@inproceedings{augenstein2020generative,
  author       = {Sean Augenstein and
                  H. Brendan McMahan and
                  Daniel Ramage and
                  Swaroop Ramaswamy and
                  Peter Kairouz and
                  Mingqing Chen and
                  Rajiv Mathews and
                  Blaise Ag{\"{u}}era y Arcas},
  title        = {Generative Models for Effective {ML} on Private, Decentralized Datasets},
  booktitle    = {International Conference on Learning Representations ({ICLR})},
  year         = {2020},
}

@inproceedings{jothiraj2023phoenix,
  author    = {Fiona Victoria Stanley Jothiraj and Afra Mashhadi},
  title     = {Phoenix: A Federated Generative Diffusion Model},
  booktitle = {Companion Proceedings of the ACM Web Conference 2024},
  series    = {WWW '24 Companion},
  year      = {2024},
  pages     = {1568-1577},
  address   = {Singapore, Singapore},
  publisher = {ACM},
  location  = {New York, NY, USA},
  month     = may,
  doi       = {10.1145/3589335.3651935},
  url       = {https://doi.org/10.1145/3589335.3651935}
}

@ARTICLE{Huang2024FLinAIGC,
  author={Huang, Xumin and Li, Peichun and Du, Hongyang and Kang, Jiawen and Niyato, Dusit and Kim, Dong In and Wu, Yuan},
  journal={IEEE Network}, 
  title={Federated Learning-Empowered AI-Generated Content in Wireless Networks}, 
  year={2024},
  volume={38},
  number={5},
  pages={304-313},
  doi={10.1109/MNET.2024.3353377}}

@INPROCEEDINGS{Hardy2019MD-GAN,
  author={Hardy, Corentin and Le Merrer, Erwan and Sericola, Bruno},
  booktitle={International Parallel and Distributed Processing Symposium (IPDPS)}, 
  title={MD-GAN: Multi-Discriminator Generative Adversarial Networks for Distributed Datasets}, 
  year={2019},
  pages={866-877},
}

@InProceedings{fan2020federated,
author="Fan, Chenyou
and Liu, Ping",
title="Federated Generative Adversarial Learning",
booktitle="Chinese Conference on Pattern Recognition and Computer Vision (PRCV)",
year="2020",
pages="3--15",
}

@ARTICLE{Li2022IFL-GAN,
  author={Li, Wei and Chen, Jinlin and Wang, Zhenyu and Shen, Zhidong and Ma, Chao and Cui, Xiaohui},
  journal={IEEE Transactions on Neural Networks and Learning Systems}, 
  title={IFL-GAN: Improved Federated Learning Generative Adversarial Network With Maximum Mean Discrepancy Model Aggregation}, 
  year={2022},
  pages={1-14},
}

@article{gupta2018distributed,
  author       = {Otkrist Gupta and
                  Ramesh Raskar},
  title        = {Distributed learning of deep neural network over multiple agents},
  journal      = {Journal of Network and Computer Applications},
  volume       = {116},
  pages        = {1--8},
  year         = {2018},
}

@article{SplitFed2022, 
title={SplitFed: When Federated Learning Meets Split Learning}, 
volume={36}, 
number={8}, 
journal={AAAI Conference on Artificial Intelligence (AAAI)}, 
author={Thapa, Chandra and Mahawaga Arachchige, Pathum Chamikara and Camtepe, Seyit and Sun, Lichao}, 
year={2022}, 
pages={8485-8493} 
}

@ARTICLE{Benshun2023SL-GAN,
  author={Yin, Benshun and Chen, Zhiyong and Tao, Meixia},
  journal={IEEE Transactions on Communications}, 
  title={Predictive GAN-Powered Multi-Objective Optimization for Hybrid Federated Split Learning}, 
  year={2023},
  volume={71},
  number={8},
  pages={4544-4560},
}

@inproceedings{Ho2020,
  author    = {Jonathan Ho and Ajay N. Jain and Pieter Abbeel},
  title     = {Denoising Diffusion Probabilistic Models},
  booktitle = {Advances in Neural Information Processing Systems 33 (NeurIPS 2020)},
  year      = {2020},
  url       = {https://proceedings.neurips.cc/paper/2020/hash/4c5bcfec8584af0d967f1ab10179ca4b-Abstract.html}
}

@article{Struppek2023,
  title = {Exploiting Cultural Biases via Homoglyphs in Text-to-Image Synthesis},
  volume = {78},
  ISSN = {1076-9757},
  url = {http://dx.doi.org/10.1613/jair.1.15388},
  DOI = {10.1613/jair.1.15388},
  journal = {Journal of Artificial Intelligence Research},
  publisher = {AI Access Foundation},
  author = {Struppek, Lukas and Hintersdorf, Dominik and Friedrich, Felix and Brack, Manuel and Schramowski, Patrick and Kersting, Kristian},
  year = {2023},
  month = dec,
  pages = {1017–1068}
}

@inproceedings{balbierer2024multivocal,
  author    = {Beatrice Balbierer and Lukas Heinlein and Domenique Zipperling and Niklas K{\"u}hl},
  title     = {A Multivocal Literature Review on Privacy and Fairness in Federated Learning},
  booktitle = {Wirtschaftsinformatik 2024 Proceedings},
  year      = {2024},
  article   = {14},
  url       = {https://aisel.aisnet.org/wi2024/14}
}

@inproceedings{saharia2022photorealistic,
  author       = {Chitwan Saharia and
                  William Chan and
                  Saurabh Saxena and
                  Lala Li and
                  Jay Whang and
                  Emily L. Denton and
                  Seyed Kamyar Seyed Ghasemipour and
                  Raphael Gontijo Lopes and
                  Burcu Karagol Ayan and
                  Tim Salimans and
                  Jonathan Ho and
                  David J. Fleet and
                  Mohammad Norouzi},
  title        = {Photorealistic Text-to-Image Diffusion Models with Deep Language Understanding},
  booktitle    = {Conference on Neural Information Processing Systems (NeurIPS)},
  year         = {2022},
}

@article{KAZEROUNI2023102846,
author = {Amirhossein Kazerouni and Ehsan Khodapanah Aghdam and Moein Heidari and Reza Azad and Mohsen Fayyaz and Ilker Hacihaliloglu and Dorit Merhof},
title = {Diffusion models in medical imaging: A comprehensive survey},
journal = {Medical Image Analysis},
volume = {88},
pages = {102846},
year = {2023},
}

@article{goodfellow20gans,
  author       = {Ian J. Goodfellow and
                  Jean Pouget{-}Abadie and
                  Mehdi Mirza and
                  Bing Xu and
                  David Warde{-}Farley and
                  Sherjil Ozair and
                  Aaron C. Courville and
                  Yoshua Bengio},
  title        = {Generative adversarial networks},
  journal      = {Communications of the ACM},
  volume       = {63},
  number       = {11},
  pages        = {139--144},
  year         = {2020},
}

@inproceedings{Kingma2014VAE,
  author = {Kingma, Diederik P. and Welling, Max},
  booktitle = {International Conference on Learning Representations (ICLR)},
  title = {{Auto-Encoding Variational Bayes}},
  year = 2014
}

@article{Ramachandranpillai2024,
  title = {Bt-GAN: Generating Fair Synthetic Healthdata via Bias-transforming Generative Adversarial Networks},
  volume = {79},
  ISSN = {1076-9757},
  url = {http://dx.doi.org/10.1613/jair.1.15317},
  DOI = {10.1613/jair.1.15317},
  journal = {Journal of Artificial Intelligence Research},
  publisher = {AI Access Foundation},
  author = {Ramachandranpillai,  Resmi and Sikder,  Md Fahim and Bergstr\"{o}m,  David and Heintz,  Fredrik},
  year = {2024},
  month = apr,
  pages = {1313–1341}
}

@inproceedings{
sun2023aligning,
title={Aligning Synthetic Medical Images with Clinical Knowledge using Human Feedback},
author={Shenghuan Sun and Gregory Goldgof and Atul Butte and Ahmed Alaa},
booktitle={Thirty-seventh Conference on Neural Information Processing Systems},
year={2023},
url={https://openreview.net/forum?id=qlnlamFQEa}
}

@inproceedings{dhariwal21beating,
  author       = {Prafulla Dhariwal and
                  Alexander Quinn Nichol},
  title        = {Diffusion Models Beat GANs on Image Synthesis},
  booktitle    = {Advances in Neural Information Processing Systems (NeurIPS)},
  pages        = {8780--8794},
  year         = {2021},
}

@inproceedings{nichol21improving,
  author       = {Alexander Quinn Nichol and
                  Prafulla Dhariwal},
  title        = {Improved Denoising Diffusion Probabilistic Models},
  booktitle    = {International Conference on Machine Learning (ICML)},
  pages        = {8162--8171},
  year         = {2021},
}

@inproceedings{song20improved,
  author       = {Yang Song and
                  Stefano Ermon},
  title        = {Improved Techniques for Training Score-Based Generative Models},
  booktitle    = {Advances in Neural Information Processing Systems (NeurIPS)},
  year         = {2020},
}

@inproceedings{ronneberger15unet,
  author       = {Olaf Ronneberger and
                  Philipp Fischer and
                  Thomas Brox},
  title        = {U-Net: Convolutional Networks for Biomedical Image Segmentation},
  booktitle    = {Medical Image Computing and Computer-Assisted Intervention (MICCAI)},
  volume       = {9351},
  pages        = {234--241},
  year         = {2015},
}

@inproceedings{
mariani2024multisource,
title={Multi-Source Diffusion Models for Simultaneous Music Generation and Separation},
author={Giorgio Mariani and Irene Tallini and Emilian Postolache and Michele Mancusi and Luca Cosmo and Emanuele Rodol{\`a}},
booktitle={International Conference on Learning Representations (ICLR)},
year={2024},
}

@misc{videoworldsimulators2024,
  title={Video generation models as world simulators},
  author={Tim Brooks and Bill Peebles and Connor Holmes and Will DePue and Yufei Guo and Li Jing and David Schnurr and Joe Taylor and Troy Luhman and Eric Luhman and Clarence Ng and Ricky Wang and Aditya Ramesh},
  year={2024},
  howpublished={\url{https://openai.com/research/video-generation-models-as-world-simulators}},
  note = {Accessed: 19-June-2024},

}

@inproceedings{
singer2023makeavideo,
title={Make-A-Video: Text-to-Video Generation without Text-Video Data},
author={Uriel Singer and Adam Polyak and Thomas Hayes and Xi Yin and Jie An and Songyang Zhang and Qiyuan Hu and Harry Yang and Oron Ashual and Oran Gafni and Devi Parikh and Sonal Gupta and Yaniv Taigman},
booktitle={International Conference on Learning Representations (ICLR)},
year={2023},
}

@InProceedings{Rombach_2022_CVPR,
    author    = {Rombach, Robin and Blattmann, Andreas and Lorenz, Dominik and Esser, Patrick and Ommer, Bj\"orn},
    title     = {High-Resolution Image Synthesis With Latent Diffusion Models},
    booktitle = {Conference on Computer Vision and Pattern Recognition (CVPR)},
    year      = {2022},
    pages     = {10684-10695},
}

@inproceedings{celeba,
  title = {{Deep Learning Face Attributes in the Wild} },
  author = {Liu, Ziwei and Luo, Ping and Wang, Xiaogang and Tang, Xiaoou},
  booktitle = {International Conference on Computer Vision (ICCV)},
  year = {2015} 
}

@inproceedings{parmar2021cleanfid,
  title={On Aliased Resizing and Surprising Subtleties in GAN Evaluation},
  author={Parmar, Gaurav and Zhang, Richard and Zhu, Jun-Yan},
  booktitle    = {Conference on Computer Vision and Pattern Recognition (CVPR)},
  year={2022},
  pages = {11400--11410},
}

@inproceedings{
bińkowski2018demystifying,
title={Demystifying {MMD} {GAN}s},
author={Mikołaj Bińkowski and Dougal J. Sutherland and Michael Arbel and Arthur Gretton},
booktitle={International Conference on Learning Representations (ICLR)},
year={2018},
}

@inproceedings{NIPS2017_8a1d6947,
 author = {Heusel, Martin and Ramsauer, Hubert and Unterthiner, Thomas and Nessler, Bernhard and Hochreiter, Sepp},
 booktitle = {Advances in Neural Information Processing Systems (NeurIPS)},
 title = {GANs Trained by a Two Time-Scale Update Rule Converge to a Local Nash Equilibrium},
 year = {2017},
 pages = {6626--6637},
}

@inproceedings{chitwan22palette,
  author       = {Chitwan Saharia and
                  William Chan and
                  Huiwen Chang and
                  Chris A. Lee and
                  Jonathan Ho and
                  Tim Salimans and
                  David J. Fleet and
                  Mohammad Norouzi},
  title        = {Palette: Image-to-Image Diffusion Models},
  booktitle    = {Special Interest Group on Computer Graphics and Interactive Techniques ({SIGGRAPH})},
  pages        = {15:1--15:10},
  year         = {2022},
}

@inproceedings{wang20attacktails,
  author       = {Hongyi Wang and
                  Kartik Sreenivasan and
                  Shashank Rajput and
                  Harit Vishwakarma and
                  Saurabh Agarwal and
                  Jy{-}yong Sohn and
                  Kangwook Lee and
                  Dimitris S. Papailiopoulos},
  title        = {Attack of the Tails: Yes, You Really Can Backdoor Federated Learning},
  booktitle    = {Advances in Neural Information Processing Systems (NeurIPS)},
  year         = {2020},
}

@inproceedings{tajalli23serversidebackdoor,
  author       = {Behrad Tajalli and
                  Oguzhan Ersoy and
                  Stjepan Picek},
  title        = {On Feasibility of Server-side Backdoor Attacks on Split Learning},
  booktitle    = {IEEE Security and Privacy Workshops (SPW)},
  pages        = {84--93},
  year         = {2023},
}

@inproceedings{struppek2022rickrolling,
author    = {Lukas Struppek and
               Dominik Hintersdorf and
               Kristian Kersting},
title     = {Rickrolling the Artist: Injecting Backdoors into Text Encoders for Text-to-Image Synthesis}, 
booktitle = {International Conference on Computer Vision ({ICCV})},
year      = {2023},
}

@inproceedings{chou23diffusionbackdoor,
  title = {How to Backdoor Diffusion Models?},
  url = {http://dx.doi.org/10.1109/CVPR52729.2023.00391},
  DOI = {10.1109/cvpr52729.2023.00391},
  booktitle = {2023 IEEE/CVF Conference on Computer Vision and Pattern Recognition (CVPR)},
  publisher = {IEEE},
  author = {Chou,  Sheng-Yen and Chen,  Pin-Yu and Ho,  Tsung-Yi},
  year = {2023},
  month = jun,
  pages = {4015–4024}
}

@article{2020t5,
  author  = {Colin Raffel and Noam Shazeer and Adam Roberts and Katherine Lee and Sharan Narang and Michael Matena and Yanqi Zhou and Wei Li and Peter J. Liu},
  title   = {Exploring the Limits of Transfer Learning with a Unified Text-to-Text Transformer},
  journal = {Journal of Machine Learning Research},
  year    = {2020},
  volume  = {21},
  number  = {140},
  pages   = {1-67},
}

@inproceedings{edward22lora,
  author       = {Edward J. Hu and
                  Yelong Shen and
                  Phillip Wallis and
                  Zeyuan Allen{-}Zhu and
                  Yuanzhi Li and
                  Shean Wang and
                  Lu Wang and
                  Weizhu Chen},
  title        = {LoRA: Low-Rank Adaptation of Large Language Models},
  booktitle    = {International Conference on Learning Representations (ICLR)},
  year         = {2022},
}

@article{dockhorn2022differentially,
title={Differentially Private Diffusion Models},
author={Tim Dockhorn and Tianshi Cao and Arash Vahdat and Karsten Kreis},
journal={Transactions on Machine Learning Research (TMLR)},
issn={2835-8856},
year={2023},
url={https://openreview.net/forum?id=ZPpQk7FJXF},
note={}
}

@article{ghalebikesabi2023differentially,
      title={Differentially Private Diffusion Models Generate Useful Synthetic Images}, 
      author={Sahra Ghalebikesabi and Leonard Berrada and Sven Gowal and Ira Ktena and Robert Stanforth and Jamie Hayes and Soham De and Samuel L. Smith and Olivia Wiles and Borja Balle},
      year={2023},
      journal={arXiv Preprint},
      eprint  = {2302.13861},
  archivePrefix= {arXiv},
}

@inproceedings{carlini23diffusion,
  author       = {Nicholas Carlini and
                  Jamie Hayes and
                  Milad Nasr and
                  Matthew Jagielski and
                  Vikash Sehwag and
                  Florian Tram{\`{e}}r and
                  Borja Balle and
                  Daphne Ippolito and
                  Eric Wallace},
  title        = {Extracting Training Data from Diffusion Models},
  booktitle    = {{USENIX} Security Symposium ({USENIX})},
  pages        = {5253--5270},
  year         = {2023},
}

@inproceedings{burg21memorization,
  author       = {Gerrit J. J. van den Burg and
                  Christopher K. I. Williams},
  title        = {On Memorization in Probabilistic Deep Generative Models},
  booktitle    = {Conference on Neural Information Processing Systems (NeurIPS)},
  pages        = {27916--27928},
  year         = {2021},
}

@article{Hintersdorf2024,
  title = {Does CLIP Know My Face?},
  volume = {80},
  ISSN = {1076-9757},
  url = {http://dx.doi.org/10.1613/jair.1.15461},
  DOI = {10.1613/jair.1.15461},
  journal = {Journal of Artificial Intelligence Research},
  publisher = {AI Access Foundation},
  author = {Hintersdorf,  Dominik and Struppek,  Lukas and Brack,  Manuel and Friedrich,  Felix and Schramowski,  Patrick and Kersting,  Kristian},
  year = {2024},
  month = jul,
  pages = {1033–1062}
}

@inproceedings{hintersdorf2024finding,
author = {Hintersdorf, Dominik and Struppek, Lukas and Kersting, Kristian and Dziedzic, Adam and Boenisch, Franziska},
title = {Finding NeMo: localizing neurons responsible for memorization in diffusion models},
year = {2024},
isbn = {9798331314385},
publisher = {Curran Associates Inc.},
address = {Red Hook, NY, USA},
booktitle = {Proceedings of the 38th International Conference on Neural Information Processing Systems},
articleno = {2800},
numpages = {43},
location = {Vancouver, BC, Canada},
series = {NIPS '24}
}

@INPROCEEDINGS{mendieta2024navigating,
  author={Mendieta, Matías and Sun, Guangyu and Chen, Chen},
  booktitle={2025 IEEE/CVF Winter Conference on Applications of Computer Vision (WACV)}, 
  title={Navigating Heterogeneity and Privacy in One-Shot Federated Learning with Diffusion Models}, 
  year={2025},
  volume={},
  number={},
  pages={2601-2610},
  doi={10.1109/WACV61041.2025.00258}}

@article{goede2024fltraining,
  author       = {Matthijs de Goede and
                  Bart Cox and
                  J{\'{e}}r{\'{e}}mie Decouchant},
  title        = {Training Diffusion Models with Federated Learning},
  journal      = {arXiv preprint},
  eprint       = {2406.12575},
  archivePrefix= {arXiv},
  year         = {2024},
}

@ARTICLE{Feng.2023IoTSL,
  author={Feng, Xingyu and Luo, Chengwen and Chen, Jiongzhang and Huang, Yijing and Zhang, Jin and Xu, Weitao and Li, Jianqiang and Leung, Victor C. M.},
  journal={IEEE Internet of Things Journal}, 
  title={IoTSL: Toward Efficient Distributed Learning for Resource-Constrained Internet of Things}, 
  year={2023},
  volume={10},
  number={11},
  pages={9892-9905},
  doi={10.1109/JIOT.2023.3235765}}

@incollection{Zhao2023,
  title = {Federated Learning Based on Diffusion Model to Cope with Non-IID Data},
  ISBN = {9789819985463},
  ISSN = {1611-3349},
  url = {http://dx.doi.org/10.1007/978-981-99-8546-3_18},
  DOI = {10.1007/978-981-99-8546-3_18},
  booktitle = {Pattern Recognition and Computer Vision},
  publisher = {Springer Nature Singapore},
  author = {Zhao,  Zhuang and Yang,  Feng and Liang,  Guirong},
  year = {2023},
  month = dec,
  pages = {220--231}
}

@article{Song2024,
  title = {Fedadkd: heterogeneous federated learning via adaptive knowledge distillation},
  volume = {27},
  ISSN = {1433-755X},
  url = {http://dx.doi.org/10.1007/s10044-024-01350-4},
  DOI = {10.1007/s10044-024-01350-4},
  number = {4},
  journal = {Pattern Analysis and Applications},
  publisher = {Springer Science and Business Media LLC},
  author = {Song,  Yalin and Liu,  Hang and Zhao,  Shuai and Jin,  Haozhe and Yu,  Junyang and Liu,  Yanhong and Zhai,  Rui and Wang,  Longge},
  year = {2024},
  month = oct 
}

@inproceedings{Lai2024,
  title = {On-demand Quantization for Green Federated Generative Diffusion in Mobile Edge Networks},
  url = {http://dx.doi.org/10.1109/ICC51166.2024.10622695},
  DOI = {10.1109/icc51166.2024.10622695},
  booktitle = {ICC 2024 - IEEE International Conference on Communications},
  publisher = {IEEE},
  author = {Lai,  Bingkun and He,  Jiayi and Kang,  Jiawen and Li,  Gaolei and Xu,  Minrui and zhang,  Tao and Xie,  Shengli},
  year = {2024},
  month = jun,
  pages = {2883–2888}
}

@inproceedings{song2020denoising,
title={Denoising Diffusion Implicit Models},
author={Jiaming Song and Chenlin Meng and Stefano Ermon},
booktitle={International Conference on Learning Representations},
year={2021},
url={https://openreview.net/forum?id=St1giarCHLP}
}

@article{Ma2024,
  title = {FedST: Federated Style Transfer Learning for Non-IID Image Segmentation},
  volume = {38},
  ISSN = {2159-5399},
  url = {http://dx.doi.org/10.1609/aaai.v38i5.28199},
  DOI = {10.1609/aaai.v38i5.28199},
  number = {5},
  journal = {Proceedings of the AAAI Conference on Artificial Intelligence},
  publisher = {Association for the Advancement of Artificial Intelligence (AAAI)},
  author = {Ma,  Boyuan and Yin,  Xiang and Tan,  Jing and Chen,  Yongfeng and Huang,  Haiyou and Wang,  Hao and Xue,  Weihua and Ban,  Xiaojuan},
  year = {2024},
  month = mar,
  pages = {4053–4061}
}

@inproceedings{zajac2023,
  title     = {Exploring Continual Learning of Diffusion Models},
  author = {Zajac,  Michal and Deja,  Kamil and Kuzina,  Anna and Tomczak,  Jakub M. and Trzcinski,  Tomasz and Shkurti,  Florian and Milos,  Piotr},
  booktitle = {Proceedings of the 4th Workshop on Continual Learning in Computer Vision (CLVision), CVPR Workshops},
  year      = {2023},
  month     = jun,
  note      = {CVPR 2023 Workshop paper}
}

@INPROCEEDINGS{Duan2023,
  author={Duan, Jing and Duan, Jie and Wan, Xuefeng and Li, Yang},
  booktitle={2023 5th International Conference on Communications, Information System and Computer Engineering (CISCE)}, 
  title={Efficient Federated Learning Method for Cloud-Edge Network Communication}, 
  year={2023},
  volume={},
  number={},
  pages={118-121},
  doi={10.1109/CISCE58541.2023.10142819}}

@article{Yao2024,
  title = {PerFedRLNAS: One-for-All Personalized Federated Neural Architecture Search},
  volume = {38},
  ISSN = {2159-5399},
  url = {http://dx.doi.org/10.1609/aaai.v38i15.29576},
  DOI = {10.1609/aaai.v38i15.29576},
  number = {15},
  journal = {Proceedings of the AAAI Conference on Artificial Intelligence},
  publisher = {Association for the Advancement of Artificial Intelligence (AAAI)},
  author = {Yao,  Dixi and Li,  Baochun},
  year = {2024},
  month = mar,
  pages = {16398–16406}
}

@inproceedings {fredriskon2014attribute,
author = {Matthew Fredrikson and Eric Lantz and Somesh Jha and Simon Lin and David Page and Thomas Ristenpart},
title = {Privacy in Pharmacogenetics: An {End-to-End} Case Study of Personalized Warfarin Dosing},
booktitle = {USENIX Security Symposium},
year = {2014},
pages = {17--32},
}

@inproceedings{mi_confidence_fredriskon,
  author    = {Matt Fredrikson and
               Somesh Jha and
               Thomas Ristenpart},
  title     = {{Model Inversion Attacks that Exploit Confidence Information and Basic Countermeasures} },
  booktitle = {Conference on Computer and Communications Security (CCS)},
  pages     = {1322--1333},
  year      = {2015},
}

@article{lecun1998mnist,
  title={The MNIST database of handwritten digits},
  author={LeCun, Yann},
  journal={http://yann.lecun.com/exdb/mnist/},
  year={1998}
}

@article{xiao2017fashion,
  title={Fashion-mnist: a novel image dataset for benchmarking machine learning algorithms},
  author={Xiao, Han and Rasul, Kashif and Vollgraf, Roland},
  journal={arXiv preprint arXiv:1708.07747},
  year={2017}
}

@inproceedings{cohen2017emnist,
  title={EMNIST: Extending MNIST to handwritten letters},
  author={Cohen, Gregory and Afshar, Saeed and Tapson, Jonathan and Van Schaik, Andre},
  booktitle={2017 international joint conference on neural networks (IJCNN)},
  pages={2921--2926},
  year={2017},
  organization={IEEE}
}

@article{clanuwat2018deep,
  title={Deep learning for classical japanese literature},
  author={Clanuwat, Tarin and Bober-Irizar, Mikel and Kitamoto, Asanobu and Lamb, Alex and Yamamoto, Kazuaki and Ha, David},
  journal={arXiv preprint arXiv:1812.01718},
  year={2018}
}

\appendix
\setcounter{table}{0}
\renewcommand{\thetable}{A.\arabic{table}}
\clearpage
\section{Computational and Communication Cost}
\label{app:cost-analysis}

We compare the per-client cost of \texttt{CollaFuse} against a federated learning (FL) baseline for diffusion models. 
A diffusion model is a noise predictor $\epsilon_\theta:\mathbb{R}^d\times[0,1]\to\mathbb{R}^d$ trained with the denoising objective
\begin{equation}
\label{eq:cfm-loss}
    \mathcal{L}(\theta)=
    \mathbb{E}_{\,t\sim\mathcal{U}[1,T],\;x_0\sim p_{\text{data}},\;\epsilon\sim\mathcal{N}(0,I)}
    \big[\,\|\epsilon_\theta(x_t,t)-\epsilon\|^2\,\big],
    \qquad x_t=\alpha_t x_0+\sigma_t\epsilon,
\end{equation}
and sampled by integrating the reverse-time SDE.

\subsection{Notation and Assumptions}
\label{app:notation}

Let $\Phi_{\text{fwd}}$ denote the FLOPs of one U-Net forward pass and $\Phi_{\text{bwd}}\approx 2\Phi_{\text{fwd}}$ those of a backward pass, so one training step costs $3\Phi_{\text{fwd}}$.
Let $T$ be the number of denoising timesteps, $t_\zeta\in[0,T]$ the cut point (server: $[t_\zeta,T]$; client: $[1,t_\zeta]$), $k$ the number of clients, $N$ the samples per client, $E$ the local epochs per round, $R$ the number of rounds, $|\theta|$ the parameter count (identical across all models), and $d$ the image (or latent) dimensionality.
The two ranges share the endpoint $t_\zeta$, matching Alg.~1 (line~7), where the client samples $t^c\sim\mathcal{U}[1,t_\zeta]$ and the server $t^s\sim\mathcal{U}[t_\zeta,T]$; this single overlapping timestep is immaterial to the cost ratios below.

We adopt the following assumptions.
\textbf{($\alpha_1$)}~\emph{Shared architecture}: $\Phi_{\text{fwd}}$ is identical across all models.
\textbf{($\alpha_2$)}~\emph{Uniform timestep sampling}: timesteps are drawn uniformly from the relevant interval.
\textbf{($\alpha_3$)}~\emph{Comparable gradient coverage}: the expected number of training steps per timestep needed to reach a target loss is comparable across schemes.

\subsection{Training Cost}
\label{app:train-cost}
Each FL client trains the full network over all timesteps $[1,T]$. With a per-client step budget of $REN$ minibatch updates, this gives
\begin{equation}
\label{eq:fl-train}
    \Phi^{\text{FL}}_{\text{train}}=3REN\,\Phi_{\text{fwd}}.
\end{equation}
A \texttt{CollaFuse} client updates only $\theta^c$ and samples its training timesteps from $[1,t_\zeta]$ rather than $[1,T]$.
We stress that this restriction alone does \emph{not} reduce training cost: the training loop (Alg.~1) still performs $REN$ minibatch updates, each a full client-network forward and backward pass, so with an unchanged step budget $\Phi^{\text{CF}}_{\text{train}}=3REN\,\Phi_{\text{fwd}}=\Phi^{\text{FL}}_{\text{train}}$.
A saving arises only if the client step budget is scaled in proportion to the share of the trajectory it must learn. Under~($\alpha_3$), reaching a target loss requires a comparable number of updates \emph{per timestep} across schemes; since the client now covers $t_\zeta$ of the $T$ timesteps, a step budget of $(t_\zeta/T)\,REN$ suffices, giving
\begin{equation}
\label{eq:cf-train}
    \Phi^{\text{CF}}_{\text{train}}=3\!\left(\tfrac{t_\zeta}{T}REN\right)\!\Phi_{\text{fwd}}
    =3REN\,\Phi_{\text{fwd}}\cdot\frac{t_\zeta}{T}.
\end{equation}
\begin{proposition}[Training cost, conditional]
\label{prop:train}
    Suppose the client's per-client step budget is scaled to $(t_\zeta/T)\,REN$ (e.g.\ by reducing rounds, local epochs, or samples accordingly). Then under
    ($\alpha_1$)--($\alpha_3$), $\Phi^{\text{CF}}_{\text{train}}/\Phi^{\text{FL}}_{\text{train}}=t_\zeta/T$.
    Without such scaling, the timestep restriction leaves the number of optimizer steps and the per-step FLOPs unchanged, and the ratio is~$1$.
\end{proposition}
\noindent This is a design statement about budget allocation, not an identity that follows from the timestep partition itself, and it is therefore weaker than the inference reduction below, which holds exactly from step counts alone (\cref{prop:inf}). For $t_\zeta=100$ and $T=1000$, allocating the client $10\%$ of the per-timestep training budget reduces client-side training FLOPs by a factor of~$10$; the remaining trajectory is learned by the shared server and
amortized across clients.

\subsection{Inference Cost}
\label{app:inf-cost}

Generating one sample requires $T$ sequential network evaluations under FL, $\Phi^{\text{FL}}_{\text{inf}}=T\,\Phi_{\text{fwd}}$.
In \texttt{CollaFuse} the client evaluates only the final $t_\zeta$ steps while the server handles the rest:
\begin{equation}
\label{eq:cf-inf}
    \Phi^{\text{CF, client}}_{\text{inf}}=t_\zeta\,\Phi_{\text{fwd}},\qquad
    \Phi^{\text{CF, server}}_{\text{inf}}=(T-t_\zeta)\,\Phi_{\text{fwd}}.
\end{equation}

\begin{proposition}[Inference cost]
\label{prop:inf}
    Under ($\alpha_1$), $\Phi^{\text{CF, client}}_{\text{inf}}/\Phi^{\text{FL}}_{\text{inf}}=t_\zeta/T$
    exactly.
    If $m$ clients share a conditioning label $y$, the server trajectory depends only on $(x_T,y)$ and can be reused, reducing the amortised server cost to $(T-t_\zeta)\Phi_{\text{fwd}}/m$ while the client cost is unchanged.
\end{proposition}

\begin{proof}
Immediate from the step counts in~\eqref{eq:cf-inf}: the client runs $t_\zeta$ steps versus $T$ under FL. 
Reuse of the server trajectory across clients sharing a label $y$ requires fixing both the initial noise $x_T$ and, for stochastic DDPM sampling, the server-side random draws.
If clients require \textit{diverse} samples, the server path cannot be reused without sacrificing sample diversity, and the amortized saving does not apply.
\end{proof}

\noindent We stress the distinction between the two reductions: the inference ratio (\cref{prop:inf}) is an exact identity that follows purely from the step counts, whereas the training ratio (\cref{prop:train}) is contingent on~($\alpha_3$).

\subsection{Training Communication Cost}
\label{app:comm-cost}

We define a \emph{round} as the outer synchronization loop shared by both schemes: within each round, every client performs $E$ local training epochs over its $N$ samples, followed by a single synchronization step with the server.

Under FedAvg, each client communicates exactly once per round, uploading its local model update and downloading the aggregated model at a cost of $2|\theta|$ scalars.
\texttt{CollaFuse} has a fundamentally different communication structure.
Because the timestep partition assigns disjoint denoising ranges to client and server, each party computes its loss independently; the client minimizes its denoising objective over $[1,t_\zeta]$ without any server involvement.
The server's training data is instead supplied by the client: at each training step the client constructs a noisy image $x_{t^s}$ at a randomly sampled server timestep $t^s\sim\mathcal{U}[t_\zeta,T]$ and passes both $x_{t^s}$ and the corresponding noise $\epsilon^s$ to the server (Alg.~1, lines~7--14), providing $2d$ scalars of training data per step---$d$ for the noisy image and $d$ for the noise target.
No gradient signal flows from server to client, so local epochs $E>1$ are feasible, in contrast to layer-wise split learning, where the server must participate in every client forward pass and $E=1$ is effectively enforced.
Over one round, the client sends $EN$ such pairs, giving
\begin{equation}
\label{eq:cf-comm}
    C^{\text{CF}}_{\text{round}}=2dEN.
\end{equation}

\begin{proposition}[Communication regime]
\label{prop:comm}
    \texttt{CollaFuse} communicates less per round than FedAvg iff $|\theta|>ENd$.
\end{proposition}

\noindent This is the \emph{large-model, modest-data} regime.
Modern image-diffusion U-Nets are large (e.g., the Stable Diffusion v1.5 U-Net has
$\sim$$10^9$ parameters) so $|\theta|$ typically lies in the $10^9$ range.
The communication balance, therefore, hinges on the local dataset size.
For a modest $N=10^4$ with $d=3\cdot32\cdot32$ and $E=1$, $ENd\approx3\times10^7$ sits well below $|\theta|$ and \texttt{CollaFuse} communicates less per round.
As $N$ grows, the margin narrows: for our CelebA clients ($N\approx5.7$--$12.4\times10^4$, $d=3\cdot64\cdot64$) the term $ENd$ reaches $\sim$$10^9$, comparable to a v1.5-scale U-Net, placing the setting near the regime boundary.
Beyond it, very large local datasets or smaller models, the inequality reverses: the compute advantages (Props.~\ref{prop:train}--\ref{prop:inf}) persist, but communication becomes the dominant per-round cost.
The didactic experiments in~\cref{app:empirical} deliberately operate in this reversed regime ($|\theta|\approx2\times10^5$) to stress-test the analysis at the opposite end of the scale spectrum.
Inference additionally transmits $\hat{x}^s_{t_\zeta}$ ($d$ scalars per sample) server-to-client.

This per-round scalar count is a simplified regime analysis.
It omits timestep indices $t_s$, label/text embeddings $y$, quantization and precision, protocol overhead, the server-to-client inference transfer, and the \textit{total} communication required to reach a target quality rather than per-round volume.
A full end-to-end communication-to-quality comparison is left to future measurement.

\subsection{Client Comparison}
\label{app:client-comparison}

\begin{table}[h]
\centering
\small
\caption{Per-client cost (FLOPs; scalars for communication).
The cut point $t_\zeta$ controls the client burden: small $t_\zeta$ offloads to the server, $t_\zeta=T$ recovers the local-only baseline.
The training ratio holds under~($\alpha_3$); the inference and communication entries are exact.}
\begin{tabular}{lccc}
\toprule
    & \textbf{Training} & \textbf{Inference/sample} & \textbf{Comm./round} \\
\midrule
    FL & $3REN\,\Phi_{\text{fwd}}$ & $T\,\Phi_{\text{fwd}}$ & $2|\theta|$ \\
    \texttt{CollaFuse} & $\tfrac{t_\zeta}{T}\,3REN\,\Phi_{\text{fwd}}$ & $t_\zeta\,\Phi_{\text{fwd}}$ & $2dEN$ \\
\midrule
    Ratio (CF/FL) & $t_\zeta/T$ & $t_\zeta/T$ & $dEN/|\theta|$ \\
\bottomrule
\end{tabular}
\label{tab:cost-summary}
\end{table}

\subsection{Quality Comparison and Analytic Cost Instantiation}
\label{app:empirical}

We instantiate the analytic cost ratios of Props.~\ref{prop:train}--\ref{prop:comm} and provide a small-scale sample-quality comparison against a FedAvg-DDPM baseline across four $28\times28$ grayscale datasets: MNIST~\citep{lecun1998mnist}, FashionMNIST~\citep{xiao2017fashion}, KMNIST~\citep{clanuwat2018deep}, and EMNIST-Letters~\citep{cohen2017emnist}.
The latter three increase visual and semantic diversity (clothing items, hiragana characters, and 26 letter classes, respectively), with EMNIST-Letters yielding the richest heterogeneity.
To mirror the non-IID conditions of the main experiments, each dataset is sorted by label and partitioned into $k=5$ contiguous bands, so every client specializes in a distinct subset of classes (similar to the main experiments).
The FLOP ratios are computed from the formulas (not independently measured); the measured component is sample quality (KID).

Both methods use an identical didactic ${\approx}200\text{k}$-parameter U-Net~($\alpha_1$), with $T=500$ timesteps, cut point $t_\zeta=100$, $R=100$ rounds, and $E=5$ local epochs per round.
Unlike the deployment-scale U-Nets discussed in~\cref{app:comm-cost} ($|\theta|\sim10^9$), this model is intentionally small ($|\theta|\approx2\times10^5$) to keep the ablation tractable; as a consequence, it sits in the communication regime where~\cref{prop:comm} predicts \texttt{CollaFuse} to be \emph{disadvantaged}, which we confirm below.
We vary the per-client dataset size $N\in\{256,1024,2048\}$ to assess scalability.
FedAvg trains the full network over $[1,T]$ and averages weights each round; \texttt{CollaFuse} trains client models over $[1,t_\zeta]$ and a shared server over $(t_\zeta,T]$ with no weight exchange.
Sample quality is measured by the Kernel Inception Distance (KID~$\downarrow$)~\citep{bińkowski2018demystifying}, which quantifies the discrepancy between the distributions of generated and held-out real images and is well suited to unpaired generative evaluation.
We note this comparison is not a controlled isolation of the timestep-split mechanism alone: \texttt{CollaFuse} differs from FedAvg in personalization, the presence of a shared server model, and the absence of weight exchange, any of which may contribute to the KID difference.
We therefore present it as a preliminary ablation rather than a definitive comparison.

\paragraph{Cost ratios.}
The per-client FLOP ratios equal the theoretical $t_\zeta/T=0.20$ exactly for inference~(\cref{prop:inf}); for training~(\cref{prop:train}) the same $0.20$ holds under the gradient-coverage assumption~($\alpha_3$).
Both ratios are independent of dataset and~$N$.
For the communication regime~(\cref{prop:comm}), the inequality reverses in this deliberately small-model setting: $|\theta|\approx2\times10^5$ while $2dEN = 2\cdot784\cdot5\cdot N\in\{2.0,\,8.0,\,16.1\}\times10^6$ for $N\in\{256,1024,2048\}$---all substantially larger than $2|\theta|\approx4.1\times10^5$.
Consequently, FedAvg communicates $5$--$39\times$ fewer scalars per round than \texttt{CollaFuse} here, confirming the predicted regime boundary at the small-model end of the spectrum complementary to the large-model regime of~\cref{app:comm-cost}.

\paragraph{Sample quality.}
\Cref{tab:empirical} shows that \texttt{CollaFuse} attains \emph{lower KID} than FedAvg in every dataset/size cell, indicating that---despite operating on only $20\%$ of the denoising trajectory locally---CollaFuse clients generate images whose distribution better matches the true data distribution.
The KID gap is largest and most clearly outside the estimator noise on EMNIST-Letters at $N=256$ ($4.43\pm0.92$ vs.\ $0.31\pm0.28$), suggesting that the richer label diversity benefits most from the shared server representation; at larger $N$ the EMNIST gap remains in the same direction but its KID standard deviation grows relative to the mean, so we read it as consistent rather than individually decisive.
Taken together, these KID results indicate that \texttt{CollaFuse} preserves---and on the distributional KID metric improves---per-client generation quality relative to the FL baseline, consistent with our findings in the main experiments.

\begin{table}[t]
\centering
\footnotesize
\setlength{\tabcolsep}{6pt}
\caption{Empirical comparison across four non-IID datasets and three per-client
data sizes ($T=500$, $t_\zeta=100$, $k=5$, $R=100$, $E=5$).
KID ($\times100$) values are mean\,$\pm$\,standard deviation across $k=5$
clients; bold denotes the better mean per row.
Client-side training and inference FLOP ratios equal $t_\zeta/T=0.20$ (training
under~($\alpha_3$), inference exactly), dataset- and $N$-independent, and are
omitted. Per-round communication: FedAvg $2|\theta|\approx4.1\times10^5$ scalars
(constant in $N$); \texttt{CollaFuse} $2dEN\in\{2.0,\,8.0,\,16.1\}\times10^6$ for
$N\in\{256,1024,2048\}$. KID standard deviations exceeding the mean (e.g.\
EMNIST-Letters, $N=1024$) reflect finite-sample estimator variance; the unbiased
KID estimator can produce slightly negative per-client values when generated
images are very close to the real data.}
\begin{tabular}{llcc}
\toprule
& & \multicolumn{2}{c}{\textbf{KID} ($\downarrow$)} \\
\cmidrule(lr){3-4}
\textbf{Dataset} & $N$ & FedAvg & \texttt{CollaFuse} \\
\midrule
\multirow{3}{*}{MNIST}
  & $256$  & $3.51\pm1.34$ & $\mathbf{1.81\pm1.58}$ \\
  & $1024$ & $2.48\pm0.86$ & $\mathbf{0.95\pm0.75}$ \\
  & $2048$ & $1.12\pm0.86$ & $\mathbf{0.54\pm0.66}$ \\
\cmidrule(lr){1-4}
\multirow{3}{*}{FashionMNIST}
  & $256$  & $6.86\pm2.22$ & $\mathbf{4.58\pm1.54}$ \\
  & $1024$ & $4.96\pm2.14$ & $\mathbf{2.45\pm1.09}$ \\
  & $2048$ & $4.24\pm1.96$ & $\mathbf{2.73\pm1.35}$ \\
\cmidrule(lr){1-4}
\multirow{3}{*}{KMNIST}
  & $256$  & $8.51\pm2.35$ & $\mathbf{3.72\pm0.59}$ \\
  & $1024$ & $4.93\pm4.00$ & $\mathbf{2.01\pm1.65}$ \\
  & $2048$ & $2.13\pm1.12$ & $\mathbf{1.09\pm0.84}$ \\
\cmidrule(lr){1-4}
\multirow{3}{*}{EMNIST-Letters}
  & $256$  & $4.43\pm0.92$ & $\mathbf{0.31\pm0.28}$ \\
  & $1024$ & $1.31\pm0.91$ & $\mathbf{0.18\pm0.41}$ \\
  & $2048$ & $0.87\pm0.56$ & $\mathbf{0.17\pm0.47}$ \\
\bottomrule
\end{tabular}

\label{tab:empirical}
\end{table}

\end{document}